\def\year{2020}\relax
\documentclass[letterpaper]{article} 

\usepackage{aaai20}  
\usepackage{times}  
\usepackage{helvet} 
\usepackage{courier}  
\usepackage[hyphens]{url}  
\usepackage{graphicx} 
\usepackage[utf8]{inputenc} 
\usepackage[T1]{fontenc}    
\usepackage{booktabs}       
\usepackage{amsfonts}       
\usepackage{nicefrac}       
\usepackage{microtype}      
\usepackage{enumitem}
\usepackage{makecell}
\usepackage{graphicx}
\usepackage{caption}
\usepackage{courier}
\usepackage{multirow}
\usepackage{color}
\usepackage{subcaption}
\usepackage{todonotes}
\usepackage{amssymb}
\usepackage{gensymb}
\usepackage{enumitem}
\usepackage{multirow}
\usepackage{algorithm}
\usepackage{amsmath,nccmath}
\usepackage{algorithmic}

\usepackage{graphicx}  
\title{Graph Few-shot Learning via Knowledge Transfer}

\author{Huaxiu Yao$^1$\thanks{Correspondence to Huaxiu Yao <huaxiuyao@psu.edu>, Ying Wei <judyweiying@tencent.com>, Zhenhui Li <JessieLi@psu.edu>}, Chuxu Zhang$^2$, Ying Wei$^3$, Meng Jiang$^2$, Suhang Wang$^1$\\ \Large{\bf Junzhou Huang$^3$, Nitesh V. Chawla$^2$, Zhenhui Li$^1$}\\$^1$Pennsylvania State University, $^2$University of Notre Dame, $^3$Tencent AI Lab\\\{huaxiuyao, szw494, JessieLi\}@psu.edu,\{czhang11, mjiang2, nchawla\}@nd.edu, \{judyweiying, joehhuang\}@tencent.com}

\begin{document}

\maketitle

\begin{abstract}
Towards the challenging problem of semi-supervised node classification, there have been extensive studies. As a frontier, Graph Neural Networks (GNNs) have aroused great interest recently, which update the representation of each node by aggregating information of its neighbors. However, most GNNs have shallow layers with a limited receptive field and may not achieve satisfactory performance especially when the number of labeled nodes is quite small. To address this challenge, we innovatively propose a graph few-shot learning (GFL) algorithm that incorporates prior knowledge learned from auxiliary graphs to improve classification accuracy on the target graph. 
Specifically, a transferable metric space characterized by a node embedding and a graph-specific prototype embedding function is shared between auxiliary graphs and the target, facilitating the transfer of structural knowledge. Extensive experiments and ablation studies on four real-world graph datasets demonstrate the effectiveness of our proposed model and the contribution of each component.
\end{abstract}

\section{Introduction}
Classifying a node (e.g., predicting interests of a user) in a graph (e.g., a social network on Facebook) in a semi-supervised manner has been challenging but imperative, inasmuch as only a small fraction of nodes have access to annotations which are usually costly. 
Drawing inspiration from traditional regularization-based~\cite{zhu2003semi} and embedding-based~\cite{perozzi2014deepwalk} approaches for graph-based semi-supervised learning, graph neural networks (GNN)~\cite{kipf2016semi,velivckovic2017graph} have attracted considerable interest and demonstrated promising performance recently.

To their essential characteristics, GNNs recursively update the feature of each node through aggregation (or message passing) of its neighbors, by which the patterns of graph topology and node features are both captured. 
Nevertheless, considering that adding more layers increases the difficulty of training and over-smoothens node features~\cite{kipf2016semi}, most of existing GNNs have shallow layers with a restricted receptive field. Therefore, GNNs are inadequate to characterize the global information, and work not that satisfactorily when the number of labeled nodes is especially small. 

Inspired by recent success of few-shot learning, from a innovative perspective, we are motivated to  \emph{leverage the knowledge learned from auxiliary graphs to improve semi-supervised node classification in the target graph of our interest.} The intuition behind lies in that auxiliary graphs and the target graph likely share local topological structures as well as class-dependent node features~\cite{shervashidze2011weisfeiler,Koutra2013deltacon}. For example, an existing social network group of co-workers at Google offer valuable clues to predict interests of users in a newly emerged social network group of co-workers at Amazon.

Yet it is even more challenging to achieve few-shot learning on graphs than on independent and identically distributed data (e.g., images) which exisiting few shot learning algorithms focus on. The two lines of recent few-shot learning works, including gradient-based methods~\cite{finn2017model,ravi2016optimization} and metric-based methods~\cite{snell2017prototypical,vinyals2016matching}, formulate the transferred knowledge as parameter initializations (or a meta-optimizer) and a metric space, respectively. None of them, however, meets the crucial prerequisite of graph few-shot learning to succeed, i.e., transferring underlying structures across graphs.

To this end, we propose a novel \textbf{G}raph \textbf{F}ew-shot \textbf{L}earning (GFL) model. 
Built upon metric-based few-shot learning, the basic idea of GFL is to learn a transferable metric space in which the label of a node is predicted as the class of the nearest prototype to the node.
The metric space is practically characterized with two embedding functions, which embed a node and the prototype of each class, respectively.
Specifically, first, GFL learns the representation of each node using a graph autoencoder whose backbone is GNNs.
Second, to better capture global information, we establish a relational structure of all examples belonging to the same class, and learn the prototype of this class by applying a prototype GNN to the relational structure.
Most importantly, both embedding functions encrypting structured knowledge are transferred from auxiliary graphs to the target one, to remedy the lack of labeled nodes. 
Besides the two node-level structures, note that we also craft the graph-level representation via a hierarchical graph representation gate, to enforce that similar graphs have similar metric spaces.

To summarize, our main contributions are: (1) to the best of our knowledge, it is the first work that resorts to knowledge transfer to improve semi-supervised node classification in graphs;
(2) we propose a novel graph few-shot learning model (GFL) to solve the problem, which simultaneouly transfers node-level and graph-level structures across graphs;
(3) comprehensive experiments on four node classification tasks empirically demonstrate the effectiveness of GFL.

\section{Related Work}
In this section, we briefly introduce the relevant research lines of our work: graph neural network and few-shot learning.
\subsubsection{Graph Neural Network} Recently, a variety of graph neural network models (GNN) have been proposed to exploit the structures underlying graphs to benefit a variety of applications~\cite{kipf2016semi,zhang2019heterogeneous,tang2019robust,huang2019online,liu2019geniepath,gao2018large}. 
There are two lines of GNN methods: non-spectral methods and spectral methods. The spectral methods mainly learn 
graph representations 
in the spectral domain~\cite{defferrard2016convolutional,henaff2015deep,bruna2013spectral,kipf2016semi}, where the learned filters are based on 
Laplacian matrices. For non-spectral methods, the basic idea is to develop an aggregator to aggregate a local set of features~\cite{velivckovic2017graph,hamilton2017inductive}. These methods have achieved great success in several graph-based tasks, such as node classification and graph classification. Thus, in this paper, we are motivated to leverage GNN as the base architecture to learn the node and graph representation.
\subsubsection{Few-Shot Learning}
Few-shot/Zero-shot learning via knowledge transfer has achieve great success in a variety of applications~\cite{li-etal-2019-transferable,li2018hierarchical,yao2019learning}. There are two popular types of approaches for few-shot learning: (1) gradient-based few-shot learning methods, which aim to learn a better initialization of model parameters that can be updated by a few gradient steps in future tasks~\cite{finn2017model,finn2018probabilistic,lee2018gradient,yao2019hierarchically} or directly use a meta-optimizer to learn the optimization process~\cite{ravi2016optimization}; (2) metric-based few-shot learning methods, which propose to learn a generalized metric and matching functions from training tasks~\cite{snell2017prototypical,vinyals2016matching,yang2018learning,bertinetto2018meta}. Our proposed GFL falls into the second category. These traditional metric-based few-shot learning methods are dedicated to independent and identically distributed data, between which no explicit interactions exists. 

\section{Preliminaries}
\label{sec:preliminaries}
\subsubsection{Graph Neural Network}
A graph \begin{small}$\mathcal{G}$\end{small} is represented as (\begin{small}$\mathbf{A}$\end{small}, \begin{small}$\mathbf{X}$\end{small}), where \begin{small}$\mathbf{A}\in\{0, 1\}^{n\times n}$\end{small} is the adjacent matrix, and \begin{small}$\mathbf{X}=\{\mathbf{x}_1,\ldots,\mathbf{x}_n\}\in\mathbb{R}^{n\times h}$\end{small} is the node feature matrix. To learn the node represetation for graph \begin{small}$\mathcal{G}$\end{small}, an embedding function \begin{small}$f$\end{small} with parameter \begin{small}$\theta$\end{small} are defined. In this work, following the ``message-passing'' architecture~\cite{gilmer2017neural}, the embedding function \begin{small}$f_{\theta}$\end{small} is built upon graph neural network (GNN) in an end-to-end manner, which is formulated as:
\begin{equation}
\small
    \mathbf{H}^{(l+1)}=\mathcal{M}(\mathbf{A},\mathbf{H}^{(l)};\mathbf{W}^{(l)}),
\end{equation}
where \begin{small}$\mathcal{M}$\end{small} is the message passing function and has a series of possible implementations~\cite{hamilton2017inductive,kipf2016semi,velivckovic2017graph}, \begin{small}$\mathbf{H}^{(l+1)}$\end{small} is the node embedding after $l$ layers of GNN and \begin{small}$\mathbf{W}^{(l)}$\end{small} is learnable weight matrix of layer \begin{small}$l$\end{small}. The node feature \begin{small}$\mathbf{X}$\end{small} is used as the initial node embedding \begin{small}$\mathbf{H}^{(1)}$\end{small}, i.e., \begin{small}$\mathbf{H}^{(1)}=\mathbf{X}$\end{small}. After stacking \begin{small}$L$\end{small} graph neural network layers, we can get the final representation \begin{small}$\mathbf{Z}=f_{\theta}(\mathbf{A},\mathbf{X})=\mathbf{H}^{(L+1)}\in \mathbb{R}^{h'}$\end{small}. For simplicity, we will use \begin{small}$\mathbf{Z}=\mathrm{GNN}(\mathbf{A}, \mathbf{X})$\end{small} to denote a GNN with \begin{small}$L$\end{small} layers.
\subsubsection{The Graph Few-Shot Learning Problem} 
Similar as the traditional few-shot learning settings~\cite{snell2017prototypical,vinyals2016matching,finn2017meta}, in graph few-shot learning, we are given a sequence of graphs \begin{small}$\{\mathcal{G}_1, \ldots, \mathcal{G}_{N_t}\}$\end{small} sampled from a probability distribution \begin{small}$\mathcal{E}$\end{small} over tasks~\cite{baxter1998theoretical}. 
For each graph \begin{small}$\mathcal{G}_i\sim \mathcal{E}$\end{small}. we are provided with a small set of $n^{s_i}$ labeled \emph{support} nodes set \begin{small}$\mathcal{S}_i=\{(\mathbf{x}_{i,j}^{s_i},y_{i,j}^{s_i})\}_{j=1}^{n^{s_i}}$\end{small} and a \emph{query} nodes set \begin{small}$\mathcal{Q}_i=\{(\mathbf{x}_{i,j}^{q_i},y_{i,j}^{q_i})\}_{j=1}^{n^{q_i}}$\end{small}, where \begin{small}$y_{i,j}\in\{1,...K\}$\end{small} is the corresponding label. For each node \begin{small}$j$\end{small} in query set \begin{small}$\mathcal{Q}_i$\end{small}, we are supposed to predict its corresponding label by associating its embedding \begin{small}$f_{\theta}(\mathbf{A}, \mathbf{x}^{q_i}_{i,j}): \mathbb{R}^{h}\rightarrow \mathbb{R}^{h'}$\end{small} with representation (\begin{small}$f_{\theta}(\mathbf{A}, \mathbf{x}_{i,j}^{s_i}),y_{i,j}^{s_i}$\end{small}) in support set \begin{small}$\mathcal{S}_i$\end{small} via the similarity measure \begin{small}$d$\end{small}.
Specifically, in prototypical network~\cite{snell2017prototypical}, the prototype \begin{small}$\mathbf{c}_i^k$\end{small} for each class \begin{small}$k$\end{small} is defined as \begin{small}$\mathbf{c}_i^k=\sum_{\mathbf{x}^{s_i}_{i,j}\in \mathcal{S}_i^k}f_{\theta}(\mathbf{A}, \mathbf{x}_{i,j}^{s_i})/|\mathcal{S}_i^k|$\end{small}, where \begin{small}$\mathcal{S}_i^k$\end{small} denotes the sample set in \begin{small}$\mathcal{S}_i$\end{small} of class \begin{small}$k$\end{small} and \begin{small}$|\mathcal{S}_i^k|$\end{small} means the number of samples in \begin{small}$\mathcal{S}_i^k$\end{small}. For each graph \begin{small}$\mathcal{G}_i$\end{small}, the effectiveness on query set \begin{small}$\mathcal{Q}_i$\end{small} is evaluated by the loss \begin{small}$\mathcal{L}_i=\sum_k\mathcal{L}_i^k$\end{small}, where:
\begin{equation}
\label{eq:protonet}
\small
    \mathcal{L}_i^k=-\sum_{(\mathbf{x}_{i,j}^{q_i},y_{i,j}^{q_i})\in\mathcal{Q}_i^k}\log \frac{\exp(-d(f_{\theta}(\mathbf{A}, \mathbf{x}_{i,j}^{q_i}),\mathbf{c}_i^k))}{\sum_{k'}\exp(-d(f_{\theta}(\mathbf{A}, \mathbf{x}_{i,j}^{q_i}),\mathbf{c}_i^{k'}))}, 
\end{equation}
where $\mathcal{Q}_i^k$ is the query set of class $k$ from  $\mathcal{Q}_i$. 
The goal of graph few-shot learning is to learn a well-generalized embedding function \begin{small}$f_\theta$\end{small} from previous graphs which can be used to a new graph with a small support set. To achieve this goal, few-shot learning often includes two-steps, i.e., meta-training and meta-testing. In meta-training, the parameter \begin{small}$\theta$\end{small} of embedding function \begin{small}$f_\theta$\end{small} is optimized to minimize the expected empirical loss over all historical training graphs, i.e., \begin{small}$\min_{\theta}\sum_{i=1}^{N_t}\mathcal{L}_i$\end{small}. Once trained, given a new graph \begin{small}$\mathcal{G}_t$\end{small}, the learned embedding function \begin{small}$f_{\theta}$\end{small} can be used to improve the learning effectiveness with a few support nodes.

\section{Methodology}
\begin{figure*}[!t]
\begin{center}
\includegraphics[scale=0.48]{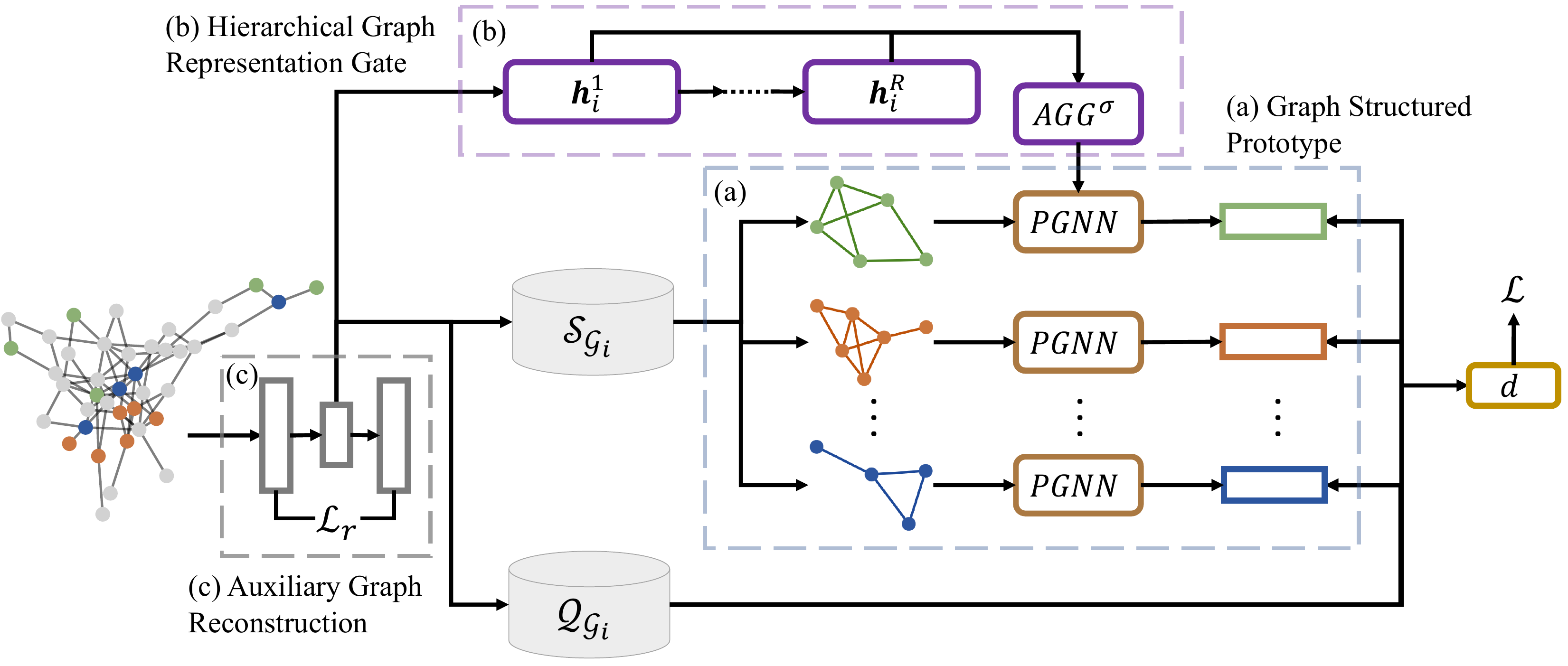}
\caption{The framework of proposed GFL including three components. (a) Graph structured prototype: we extract the graph relation structure for the support set $\mathcal{S}_i^k$ of each class $k$ and use a prototype GNN (PGNN) to learn its representation $\mathbf{c}^k_i$; (b) Hierarchical graph representation gate: we learn the hierarchical representations of graph from level 1 to R, i.e., $\mathbf{h}_i^1$ to $\mathbf{h}_i^R$, and use the aggregated representation to modulate the parameters of PGNN; (c) Auxiliary graph reconstruction: we construct the graph autoencoder to improve the training stability and the quality of node representation.} 
\label{fig:framwork}
\end{center}
\end{figure*}

In this section, we elaborate our proposed GFL whose framework is illustrated in Figure~\ref{fig:framwork}. The goal of GFL is to adapt graph-structured knowledge learned from existing graphs to the new graph \begin{small}$\mathcal{G}_t$\end{small} by exploiting the relational structure in both node-level and graph-level. In the node level, GFL captures the relational structure among different nodes. In part (a) of Figure~\ref{fig:framwork}, for each class $k$, the corresponding relational structure is constructed by the samples in \begin{small}$\mathcal{S}_i^k$\end{small} and a prototype GNN (PGNN) is proposed to learn its prototype representation. For graph-level structures, GFL learns the representation of a whole graph and ensures similar graphs have similar structured knowledge to be transferred. Specifically, as illustrated in part (b) (see Figure~\ref{fig:framwork_b} for more detailed structure of part (b)), the parameters of PGNN highly depend on the whole graph structure which is represented in a hierarchical way. The matching loss is finally computed via the similarity measure \begin{small}$d$\end{small}. To enhance the stability of training and the quality of node representation, we further introduce the auxiliary graph reconstruction structure, i.e., the part (c). In the remaining of this section, we will detail the three components, i.e., \emph{graph structured prototype}, \emph{hierarchical graph representation gate} and \emph{auxiliary graph reconstruction}.
\subsection{Graph Structured Prototype}
\label{sec:graph_proto}
In most of the cases, a node plays two important roles in a graph: one is locally interacting with the neighbors that may belong to different classes; the other is interacting with the nodes of the same class in relatively long distance, which can be globally observed. For example, on a biomedical knowledge graph, it is necessary to model both (1) the local structure among disease nodes, treatment nodes, gene nodes, and chemical nodes, and (2) the structure between disease nodes describing their co-occurrence or evolutionary relationship, as well as the structure between gene nodes describing their co-expression. The embedding results \begin{small}$\mathbf{Z}_i$\end{small} of graph \begin{small}$\mathcal{G}_i$\end{small} describes the first role of local heterogeneous information. And we need to learn the prototype of each class (as defined in Section 3) for the second role of global homogeneous information. It is non-trivial to model the relational structure among support nodes and learn their corresponding prototype, we thus propose a prototype GNN model denoted as PGNN to tackle this challenge.

Given the representation of each node, we first extract the relational structure of samples belong to class \begin{small}$k$\end{small}. For each graph \begin{small}$\mathcal{G}_i$\end{small}, the relational structure \begin{small}$\mathcal{R}_i^k$\end{small} of the sample set \begin{small}$\mathcal{S}_i^k$\end{small} can be constructed based on some similarity metrics, such as the number of k-hop common neighbors, the inverse topological distance between nodes. To improve the robustness of \begin{small}$\mathcal{R}_i^k$\end{small} and alleviate the effect of outlier nodes, we introduce a threshold \begin{small}$\mu$\end{small}. If the similarity score $w$ between a pair of nodes is smaller than \begin{small}$\mu$\end{small}, we set it to a fixed value \begin{small}$\mu_{0}$\end{small}, i.e., \begin{small}$w = \mu_{0}$\end{small} (\begin{small}$\mu_{0} < \mu$\end{small}). Then, the PGNN is used to model the interactions between samples in the \begin{small}$\mathcal{S}_i^k$\end{small}, i.e.,  \begin{small}$\mathrm{PGNN}_\phi(\mathcal{R}_i^k,f_{\theta}(\mathcal{S}_i^k))$\end{small}, where PGNN is parameterized by \begin{small}$\phi$\end{small}. Note that, \begin{small}$\mathrm{PGNN}_\phi(\mathcal{R}_i^k,f_{\theta}(\mathcal{S}_i^k))$\end{small} is a representation matrix, and we use \begin{small}$j$\end{small} to indicate the \begin{small}$j$\end{small}-th node representation. Thus, the graph structured prototype is calculated as follows,
\begin{equation}
\label{eq:graph_proto}
\small
    \mathbf{c}_i^k=\mathrm{Pool}_{j=1}^{n^{s_i^k}}(\mathrm{PGNN}_{\phi}(\mathcal{R}_i^k, f_{\theta}(\mathcal{S}_i^k))[j]),
\end{equation}
where \begin{small}$\mathrm{Pool}$\end{small} operator denotes a max or mean pooling operator over support nodes and \begin{small}$n^{s_i^k}$\end{small} represents the number of nodes in support set \begin{small}$\mathcal{S}_i^k$\end{small}.
\subsection{Hierarchical Graph Representation Gate}
\label{sec:gate}
The above prototype construction process is highly determined by the PGNN with the globally shared parameter $\phi$. However, different graphs have their own topological structure, motivating us to tailor the globally shared information to each graph. Thus, we learn a hierarchical graph representation for extracting graph-specific information and incorporate it with the parameter of PGNN through a gate function. Before detailing the structure, we first illustrate the importance of the hierarchical graph representation: the simple, node level representation for a graph can be insufficient for many complex graph structures, especially those high-order structures that widely exist and are valuable in real-world applications~\cite{morris2019weisfeiler}. 

\begin{figure}[h]
\begin{center}
\includegraphics[scale=0.26]{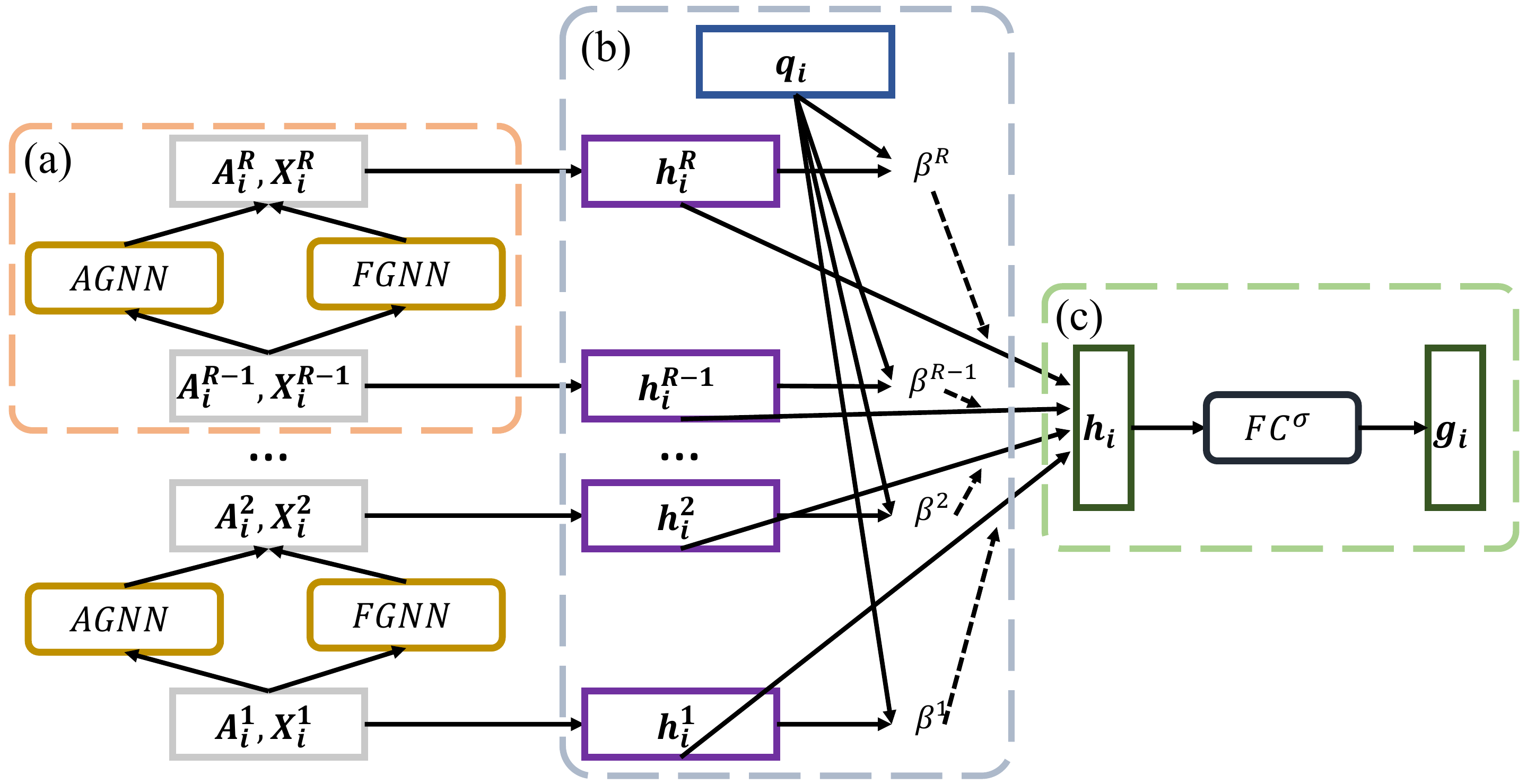}
\caption{The detailed framework of hierarchical graph representation gate: (a) the basic block for learning hierarchical representation $\{\mathbf{h}_i^1,\ldots,\mathbf{h}_i^R\}$; (b) aggregator to aggregate $\{\mathbf{h}_i^1,\ldots,\mathbf{h}_i^R\}$, where the learnable query vector $\mathbf{q}_i$ is introduced to calculate the attention weight $\beta^1\cdots\beta^R$. Note that, we only illustate the attention aggregator. 
(c) graph-specific gate construction, where $\mathbf{h}_i$ is used to calculate gate $\mathbf{g}_i$.} 
\label{fig:framwork_b}
\end{center}
\end{figure}

Figure~\ref{fig:framwork_b} illustrates the detailed structure of hierarchical graph representation gate. First, following the popular method of hierarchical graph modeling~\cite{ying2018hierarchical}, the hierarchical graph representation for each level is accomplished by alternating between two level-wise stages: the node assignment and the representation fusion (see part (a) in Figure~\ref{fig:framwork_b}). 
\\
\textbf{Node Assignment}
In the assignment step, each low-level node is assigned to high-level community. In level \begin{small}$r$\end{small} of graph \begin{small}$\mathcal{G}_i$\end{small}, we denote the number of nodes as \begin{small}$K^{r}$\end{small}, the adjacency matrix as \begin{small}$\mathbf{A}_i^r$\end{small}, the feature matrix as \begin{small}$\mathbf{X}_i^r$\end{small}. For the \begin{small}$k^{r}$\end{small}-th node in \begin{small}$r$\end{small}-th level, the assignment value \begin{small}$p_i^{k^{r}\rightarrow k^{r+1}}$\end{small} from node \begin{small}$k^{r}$\end{small} to node \begin{small}$k^{r+1}$\end{small} in \begin{small}$(r+1)$\end{small}-th level is calculated by applying softmax function on the output of an assignment GNN (AGNN) as follows,
\begin{equation}
\label{eq:agnn}
\small
    p_i^{k^{r}\rightarrow k^{r+1}}=\frac{\exp(\mathrm{AGNN}(\mathbf{A}_i^{r}, \mathbf{X}_i^{r})[k^{r},k^{r+1}])}{\sum_{\bar{k}^{r+1}=1}^{K^{r+1}}\exp(\mathrm{AGNN}(\mathbf{A}_i^{r}, \mathbf{X}_i^{r})[k^{r},\bar{k}^{r+1}])},
\end{equation}
where 
\begin{small}$\mathrm{AGNN}(\mathbf{A}_i^{r}, \mathbf{X}_i^{r})[k^{r},k^{r+1}]\in \mathbb{R}^1$\end{small} denotes the assignment representation value from node \begin{small}$k^{r}$\end{small} in level \begin{small}$r$\end{small} to node \begin{small}$k^{r+1}$\end{small} in level \begin{small}$r+1$\end{small}. The whole assignment matrix including every assignment probability \begin{small}$p_i^{k^{r}\rightarrow k^{r+1}}$\end{small} is denoted as \begin{small}$\mathbf{P}_{i}^{r\rightarrow r+1}\in \mathbb{R}^{K^r\times K^{r+1}}$\end{small}.
\\
\textbf{Representation Fusion}
After getting the assignment matrix \begin{small}$\mathbf{P}_{i}^{r\rightarrow r+1}\in \mathbb{R}^{K^r\times K^{r+1}}$\end{small}, for level \begin{small}$r+1$\end{small}, the adjacent matrix is defined as \begin{small}$\mathbf{A}_i^{r+1}=(\mathbf{P}_{i}^{r\rightarrow r+1})^T\mathbf{A}_i^r\mathbf{P}_{i}^{r\rightarrow r+1}$\end{small} and the feature matrix is calculated by applying assignment matrix on the output of a fusion GNN (FGNN), i.e., \begin{small}$\mathbf{X}_i^{r+1}=(\mathbf{P}_{i}^{r\rightarrow r+1})^T\mathrm{FGNN}(\mathbf{A}_i^r, \mathbf{X}_i^r)$\end{small}. Then, the feature representation \begin{small}$ \mathbf{h}_i^{r+1}$\end{small} of level \begin{small}$r+1$\end{small} can be calculated by aggregating the representation of all nodes, i.e.,
\begin{equation}
\label{eq:graph_repr}
\small
    \mathbf{h}_i^{r+1}=\mathrm{Pool}_{k^{r+1}=1}^{K^{r+1}}((\mathbf{P}_{i}^{r\rightarrow r+1})^T\mathrm{FGNN}(\mathbf{A}_i^r, \mathbf{X}_i^r)[k^{r+1}]),
\end{equation}
where \begin{small}$\mathbf{X}_i^{r+1}[k^{r+1}]=(\mathbf{P}_{i}^{r\rightarrow r+1})^T\mathrm{FGNN}(\mathbf{A}_i^r, \mathbf{X}_i^r)[k^{r+1}]$\end{small} denotes the feature representation of node \begin{small}$k^{r+1}$\end{small}. 

By calculating the representation of each level, we get the representation set \begin{small}$\{\mathbf{h}_i^1,\ldots, \mathbf{h}_i^R\}$\end{small} which encrypts the graph structure from different levels. 
Then, to get the whole graph representation \begin{small}$\mathbf{h}_i$\end{small},
the representation of each level is aggregated via an aggregator \begin{small}$\mathrm{AGG}$\end{small}. In this work, we propose two candidate aggregators: mean pooling aggregator and attention aggregator (see part (b) in Figure~\ref{fig:framwork_b}). For mean pooling aggregator, the graph representation \begin{small}$\mathbf{h}_i$\end{small} is defined as:
\begin{equation}
\label{eq:graph_repr_mean}
\small
    \mathbf{h}_i=\mathrm{AGG}_{\mathrm{mean}}(\{\mathbf{h}_i^1,\ldots, \mathbf{h}_i^R\})=\frac{1}{R}\sum_{r=1}^{R}\mathbf{h}_i^r.
\end{equation}
Considering the representation of different levels may have different contributions on the whole representation \begin{small}$\mathbf{h}_i$\end{small}, for attention aggregator, we first introduce a learnable query vector as \begin{small}$\mathbf{q}_i$\end{small}, and then the formulation is
\begin{equation}
\label{eq:graph_repr_att}
\small
    \mathbf{h}_i=\mathrm{AGG}_{\mathrm{att}}(\{\mathbf{h}_i^1,\ldots, \mathbf{h}_i^R\})=\sum_{r=1}^{R}\beta^r_i\mathbf{h}_i^r=\sum_{r=1}^{R}\frac{\mathbf{q}_i^T\mathbf{h}_i^r}{\sum_{r^{'}=1}^{R}\mathbf{q}_i^T\mathbf{h}_i^{r^{'}}}\mathbf{h}_i^r.
\end{equation}
After the aggregating process, the final representation \begin{small}$\mathbf{h}_i$\end{small} is expected to be graph-specific. Inspired by previous findings~\cite{xu2015show}: similar graphs may activate similar parameters (i.e., parameter \begin{small}$\phi$\end{small} of the PGNN), we introduce a gate function $\mathbf{g}_i=\mathcal{T}(\mathbf{h}_i)$ (see part (c) in Figure~\ref{fig:framwork_b}) to tailor graph structure specific information. Then, the global transferable knowledge (i.e., $\phi$) is adapted to the structure-specific parameter via the gate function, which is defined as follows:
\begin{equation}
\small
\label{eq:gated_proto}
    \phi_i=\mathbf{g}_i\circ \phi=\mathcal{T}(\mathbf{h}_i)\circ \phi,
\end{equation}
where $\circ$ represents element-wise multiplication. \begin{small}$\mathbf{g}_i=\mathcal{T}(\mathbf{h}_i)$\end{small} maps the graph-specific representation \begin{small}$\mathbf{h}_i$\end{small} to the same space of parameter $\phi$, which is defined as:
\begin{equation}
\label{eq:gate}
\small
    \mathbf{g}_i=\mathcal{T}(\mathbf{h}_i)=\sigma(\mathbf{W}_g\mathbf{h}_i+\mathbf{b}_g),
\end{equation}
where \begin{small}$\mathbf{W}_g$\end{small} and \begin{small}$\mathbf{b}_g$\end{small} are learnable parameters. Thus, \begin{small}$\mathrm{PGNN}_{\phi}$\end{small} in Eqn.~\eqref{eq:graph_proto} would be \begin{small}$\mathrm{PGNN}_{\phi_i}$\end{small}.

\subsection{Auxiliary Graph Reconstruction}
In practice, it is difficult to learn an informative node representation using only the signal from the matching loss, which motivates us to design a new constraint for improving the training stability and the quality of node representation. Thus, for the node embedding function $f_{\theta}(\cdot)$, we refine it by using a graph autoencoder. The reconstruction loss for training autoencoder is defined as follows,
\begin{equation}
\label{eq:reconstruction_loss}
\small
    \mathcal{L}_r(\mathbf{A}_i, \mathbf{X}_i)=\Vert \mathbf{A}_i-\mathrm{GNN}_{dec}(\mathbf{Z}_i)\mathrm{GNN}_{dec}^T(\mathbf{Z}_i)\Vert_F^2,
\end{equation}
where \begin{small}$\mathbf{Z}_i=\mathrm{GNN}_{enc}(\mathbf{A}_i,\mathbf{H}_i)$\end{small} is the representation for each node in graph \begin{small}$\mathcal{G}_i$\end{small} and \begin{small}$\Vert\cdot\Vert_F$\end{small} represents the Frobenius norm. Recalling the objectives for a prototypical network in Section~\ref{sec:preliminaries}, we reach the optimization problem of GFL as \begin{small}$\min_{\Theta}\sum_{i=1}^{N_t}\mathcal{L}_i+\gamma\mathcal{L}_r(\mathbf{A}_i, \mathbf{X}_i)$\end{small}, where \begin{small}$\Theta$\end{small} represents all learnable parameters. 
The whole training process of GFL is detailed in Alg.~\ref{alg:gfl}.
\begin{algorithm}[tb]
    \caption{Training Process of GFL}
    \label{alg:gfl}
    \begin{algorithmic}[1]
    \REQUIRE \begin{small}$\mathcal{E}$\end{small}: distribution over graphs; \begin{small}$L$\end{small}: \# of layers in hierarchical structure; \begin{small}$\alpha$\end{small}: stepsize; \begin{small}$\gamma$\end{small}: balancing parameter for loss
    \STATE Randomly initialize \begin{small}$\Theta$\end{small}
    \WHILE{not done}
    \STATE Sample a batch of graphs \begin{small}$\mathcal{G}_i\sim\mathcal{E}$\end{small} and its corresponding adjacent matrices \begin{small}$\mathbf{A}_i$\end{small} and feature matrices \begin{small}$\mathbf{X}_i$\end{small}
    \FORALL{\begin{small}$\mathcal{G}_i$\end{small}}
    \STATE Sample support set \begin{small}$\mathcal{S}_i$\end{small} and query set \begin{small}$\mathcal{Q}_i$\end{small}
    \STATE Compute the embedding \begin{small}$f_{\theta}(\mathbf{A}_i, \mathbf{X}_i)$\end{small} and its reconstruction error \begin{small}$\mathcal{L}_r(\mathbf{A}_i, \mathbf{X}_i)$\end{small} in Eqn.~\eqref{eq:reconstruction_loss}
    \STATE Compute the hierarchical representation \begin{small}$\{\mathbf{h}_i^1,\ldots, \mathbf{h}_i^R\}$\end{small} in Eqn.~\eqref{eq:graph_repr} and gated parameter \begin{small}$\phi_i$\end{small} in Eqn.~\eqref{eq:gated_proto}
    \STATE Construct relational graphs \begin{small}$\{\mathcal{R}_i^1,...,\mathcal{R}_i^K\}$\end{small} for samples in \begin{small}$\mathcal{S}_i$\end{small} and compute graph prototype \begin{small}$\{\mathbf{c}_i^1,...,\mathbf{c}_i^K\}$\end{small} in Eqn.~\eqref{eq:graph_proto}.
    \STATE Compute the matching score using the query set \begin{small}$\mathcal{Q}_i$\end{small} and evaluate loss in Eqn.~\eqref{eq:protonet}
    \ENDFOR
    \STATE Update \begin{small}$\Theta\leftarrow \Theta-\alpha\nabla_{\Theta} \sum_{i=1}^{N_t} \mathcal{L}_i(\mathbf{A}_i, \mathbf{X}_i)+\gamma\mathcal{L}_r(\mathbf{A}_i, \mathbf{X}_i) $\end{small}
    \ENDWHILE
    \end{algorithmic}
\end{algorithm}

\section{Experiments}
In this section, we conduct extensive experiments to demonstrate the benefits of GFL, with the goal of answering the following questions:
\begin{table}[h]
\small
\begin{center}
\caption{Data Statistics.}
\label{tab:data_statistics}
\begin{tabular}{l|cccc}
\toprule
Dataset & Colla. & Reddit & Cita. & Pubmed\\
\midrule
\# Nodes (avg.) &4,496&5,469 &2,528 & 2,901\\
\# Edges (avg.) &14,562 & 7,325&14,710 & 5,199\\
\# Features/Node & 128 & 600 & 100 & 500\\
\# Classes & 4 & 5 & 3 & 3\\
\# Graphs (Meta-train) & 100 & 150 & 30 & 60\\
\# Graphs (Meta-val.) & 10 & 15 & 3 & 5\\
\# Graphs (Meta-test) & 20 & 25 & 10 & 15\\
\midrule
\end{tabular}
\end{center}
\end{table}%
(1) Can GFL outperform baseline methods? (2) Can our proposed graph structured prototype and hierarchical graph representation gate improve the performance? (3) Can our approach learn better representations for each class?
\subsubsection{Dataset Description}
We use four datasets of different kinds of graphs: Collaboration, Reddit, Citation and Pubmed. 
(1): \emph{Collaboration data}: Our first task is to predict research domains of different academic authors. We use the collaboration graphs extracted from the AMiner data \cite{AMiner}. Each author is assigned with a computer science category label according to the majority of their papers' categories. (2): \emph{Reddit data}: In the second task, we predict communities of different Reddit posts. We construct post-to-post graphs from Reddit community data \cite{hamilton2017inductive}, where each edge denotes that the same user comments on both posts. Each post is labeled with a community id. (3): \emph{Citation data}: The third task is to predict paper categories. We derive paper citation graphs from the AMiner data and each paper is labeled with a computer science category label. (4): \emph{Pubmed data}: Similar to the third task, the last task is to predict paper class labels. The difference is that the citation graphs are extracted from the PubMed database \cite{velivckovic2017graph} and each node is associated with diabetes class id. The statistics of these datasets are reported in Table~\ref{tab:data_statistics}. 
\begin{table*}[h]
\begin{center}
\small
\caption{Comparison between GFL and other node classification methods on four graph datasets. Performance of Accuracy$\pm95\%$ confidence intervals on 10-shot classification are reported.} 
\label{tab:model_compare}
\begin{tabular}{l|c|c|c|c}
\toprule
Model & Collaboration & Reddit & Citation & Pubmed\\
\midrule
LP~\cite{zhu2002learning} & $61.09\pm1.36\%$ & $23.40\pm1.63\%$ & $67.00\pm4.50\%$ & $48.55\pm6.01\%$ \\
Planetoid~\cite{yang2016revisiting} & $62.95\pm 1.23\%$ & $50.97\pm3.81\%$ & $61.94\pm2.14\%$ & $51.43\pm3.98\%$\\
\midrule
Deepwalk~\cite{perozzi2014deepwalk} &$51.74\pm 1.59\%$  &$34.81\pm 2.81\%$ &$56.56\pm 5.25\%$ &$44.33\pm 4.88\%$ \\
node2vec~\cite{grover2016node2vec} & $59.77\pm 1.67\%$ & $43.57\pm 2.23\%$&$54.66\pm 5.16\%$ &$41.89\pm 4.83\%$\\
Non-transfer-GCN~\cite{kipf2016semi} & $63.16\pm1.47\%$ & $46.21\pm1.43\%$ & $63.95\pm5.93\%$ & $54.87\pm 3.60\%$\\
\midrule
All-Graph-Finetune (AGF) & $76.09\pm 0.56\%$ & $54.13\pm0.57\%$ & $88.93\pm0.72\%$  & $83.06\pm0.72\%$\\
K-NN & $67.53\pm 1.33\%$ & $56.06\pm1.36\%$ & $78.18\pm1.70\%$  & $74.33\pm0.52\%$\\
Matchingnet~\cite{vinyals2016matching}& $80.87\pm0.76\%$ & $56.21\pm1.87\%$ & $94.38\pm0.45\%$ & $85.65\pm0.21\%$ \\
MAML~\cite{finn2017model} & $79.37\pm0.41\%$ & $59.39\pm0.28\%$ & $95.71\pm 0.23\%$ & $88.44\pm0.46\%$\\
Protonet~\cite{snell2017prototypical}  & $80.49\pm 0.55\%$ & $60.46\pm0.67\%$ & $95.12\pm 0.17\%$ & $87.90\pm0.54\%$\\
  \midrule
\textbf{GFL-mean (Ours)} & $83.51\pm 0.38\%$ & $62.66\pm0.57\%$ & $\mathbf{96.51\pm0.31\%}$ & $\mathbf{89.37\pm0.41\%}$\\
\textbf{GFL-att (Ours)} & $\mathbf{83.79\pm 0.39\%}$ & $\mathbf{63.14\pm0.51\%}$ & $95.85\pm0.26\%$ & $88.96\pm0.43\%$\\
  \midrule
\end{tabular}
\end{center}
\end{table*}%

\subsubsection{Experimental Settings}
In this work, we follow the traditional few-shot learning settings~\cite{finn2017model,snell2017prototypical}. For each graph, $N$ labeled nodes for each class are provided as support set. The rest nodes are used as query set for evaluating the performance. Like~\cite{kipf2016semi}, the embedding structure (i.e., $\theta$ in Eqn.~\eqref{eq:graph_proto}) is a two-layer graph convolutional structure (GCN) with 32 neurons in each layer. For PGNN in Eqn.~\eqref{eq:graph_proto}, each AGNN in Eqn.~\eqref{eq:agnn} and each FGNN in Eqn.~\eqref{eq:graph_repr}, we use one-layer GCN as the proxy of GNN. The distance metric $d$ is defined as the inner product distance. For our proposed GFL, we use GFL-mean and GFL-att to represent the type of hierarchical representation aggregator (i.e., GFL-mean represents mean pooling aggregator in Eqn.~\eqref{eq:graph_repr_mean} and GFL-att represents attention aggregator in Eqn.~\eqref{eq:graph_repr_att}). The threshold $\mu_0$ in Section 4.1 for constructing relation structure of support set is set as \begin{small}$0.5$\end{small}. Detailed hyperparameter
settings can be found in Appendix~\ref{app:hyperparameter}.
\subsubsection{Baseline Methods}
For performance comparison of node classification, we consider three types of baselines: (1) \emph{Graph-based semi-supervised methods} including Label Propagation (LP)~\cite{zhu2002learning} and Planetoid~\cite{yang2016revisiting}; (2) \emph{Graph representation learning methods} including Deepwalk~\cite{perozzi2014deepwalk}, node2vec~\cite{grover2016node2vec}, Non-transfer-GCN~\cite{kipf2016semi}. Note that, for Non-transfer-GCN, we train GCN on each meta-testing graph with limited labeled data rather than transferring knowledge from meta-training graphs; (3) \emph{Transfer/few-shot methods} including All-Graph-Finetune (AGF), K-nearest-neighbor (K-NN), Matching Network (Matchingnet)~\cite{vinyals2016matching}, MAML~\cite{finn2017model}, Prototypical Network (Protonet)~\cite{snell2017prototypical}. Note that, for All-Graph-Finetune and K-NN methods, follow the settings of~\cite{triantafillou2019meta}, we first learn the parameters of the embedding structure by feeding all meta-graphs one by one. Then, we finetune the parameters or use K-NN to classify nodes based on the learned parameters of embedding structure. Each transfer/few-shot learning method uses the same embedding structure (i.e., two layers GCN) as GFL. More detailed descriptions and implementation of baselines can be found in Appendix~\ref{app:baseline}.
\subsection{Results}
\subsubsection{Overall Results}
For each dataset, we report the averaged accuracy with 95\% confidence interval over meta-testing graphs of 10-shot node classification in Table~\ref{tab:model_compare}. Comparing with graph-based semi-supervised methods and graph representation learning methods, first, we can see that transfer/few-shot methods (i.e., AGF, K-NN, Matchingnet, MAML, Protonet, GFL) significantly improve the performance, showing the power of knowledge transfer from previous learned graphs. 
Second, both GFL-mean and GFL-att achieve the best performance than other transfer/few-shot methods on four datasets, indicating the effectiveness by incorporating graph prototype and hierarchical graph representation. In addition, as a metric distance based meta-learning algorithm, GFL not only outperforms other algorithms from this research line (i.e., Matchingnet, Protonet), but also achieves better performance than MAML, a representative gradient-based meta-learning algorithm.
\subsubsection{Ablation Studies}
Since GFL integrates three essential components (i.e., graph structured prototype, hierarchical graph representation gate, auxiliary graph reconstruction), we conduct extensive ablation studies to understand the contribution of each component. Table~\ref{tab:ablation} shows the results of ablation studies on each dataset, where the best results among GFL-att and GFL-mean are reported as GFL results. Performance of accuracy are reported in this table. For the graph structured prototype, in (M1a), we first report the performance of protonet for comparison since Protonet use mean pooling of node embedding instead of constructing and exploiting relational structure for each class.

To show the effectiveness of hierarchical graph representation gate, we first remove this component and report the performance in (M2a). The results are inferior, demonstrating that the effectiveness of graph-level representation. In addition, we only use the flat representation structure (i.e., \begin{small}$R=1$\end{small}) in (M2b). The results show the effectiveness of hierarchical representation.

For auxiliary graph reconstruction, we remove the decoder GNN and only use the encoder GNN to learn the node representation in (M3). GFL outperforms (M3) as the graph reconstruction loss refines the learned node representation and enhance the stability of training.

\begin{table*}[h]
\begin{center}
\small
\caption{Results of Ablation Study. Performance of Accuracy$\pm95\%$ confidence intervals are reported. We select the best performance of GFL-mean and GFL-att as GFL in this table.} 
\label{tab:ablation}
\begin{tabular}{p{190pt}|c|c|c|c}
\toprule
Ablation Model & Collaboration & Reddit & Citation & Pubmed\\
\midrule
(M1a): use the mean pooling prototype (i.e., protonet) & $80.49\pm 0.55\%$ & $60.46\pm0.67\%$ & $95.12\pm0.17\%$ & $87.90\pm0.54\%$\\\midrule
(M2a): remove the hierarchical representation gate & $82.63\pm 0.45\%$ & $61.99\pm 0.27\%$ & $95.33\pm0.35\%$ & $88.15\pm 0.55\%$\\
(M2b): use flat representation rather than hierarchical & $83.45\pm0.41\%$ & $62.55\pm 0.65\%$ & $95.76\pm 0.37\%$ & $89.08\pm 0.47\%$\\\midrule
(M3): remove the graph reconstruction loss & $82.98\pm0.37\%$ & $62.58\pm0.47\%$ & $95.63\pm0.27\%$ & $89.11\pm0.43\%$ \\\midrule
  \textbf{GFL (Ours)} & $\mathbf{83.79\pm 0.39\%}$ & $\mathbf{63.14\pm0.51\%}$ & $\mathbf{96.51\pm 0.31\%}$ & $\mathbf{89.37\pm0.41\%}$\\\bottomrule
\end{tabular}
\end{center}
\end{table*}
\begin{figure}[!t]
	\centering
	\begin{subfigure}[b]{0.22\textwidth}
		\centering
		\includegraphics[height=0.8\textwidth]{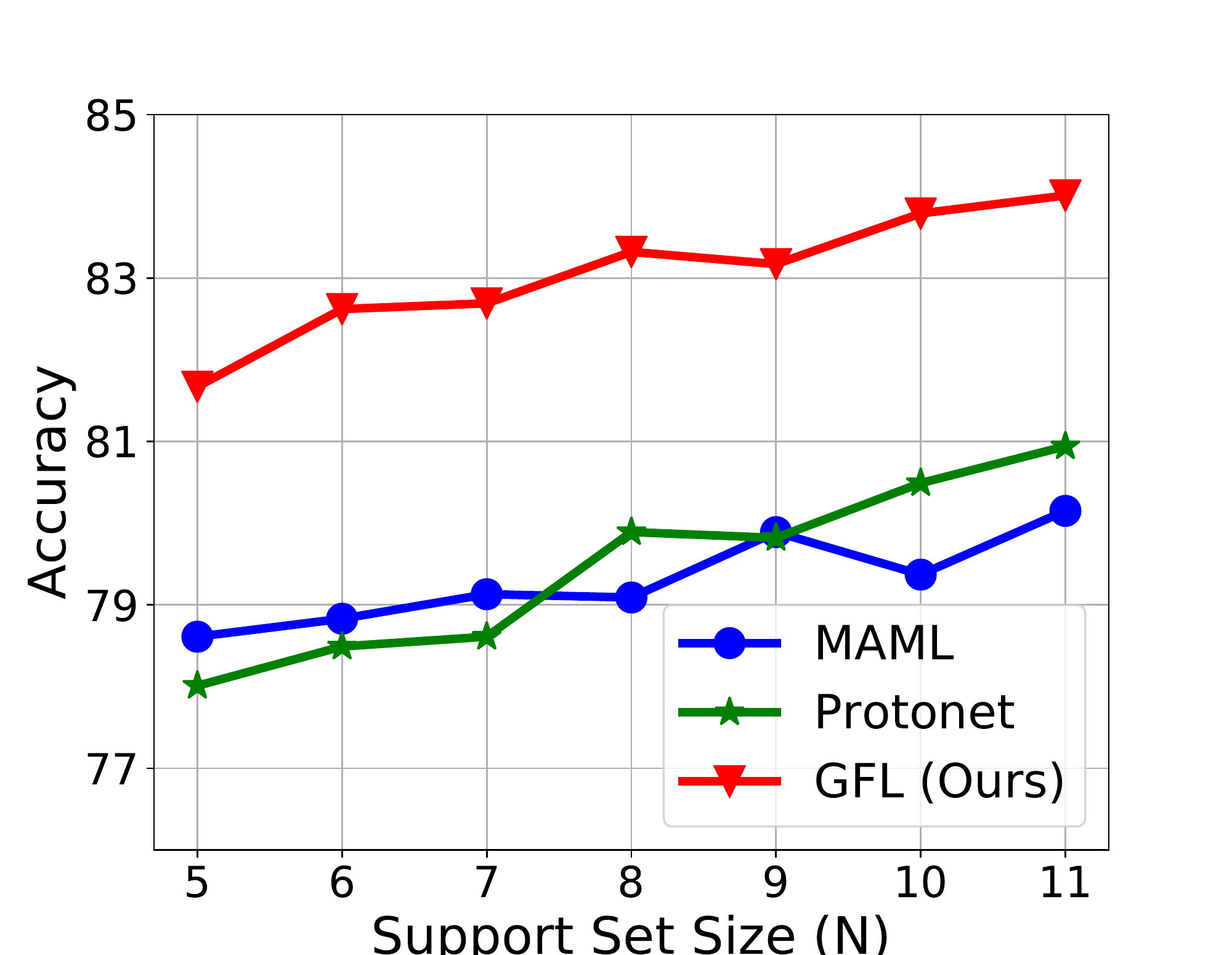}
		\caption{\label{fig:collaboration_shot}: Collaboration}
	\end{subfigure}
	\begin{subfigure}[b]{0.22\textwidth}
		\centering
		\includegraphics[height=0.8\textwidth]{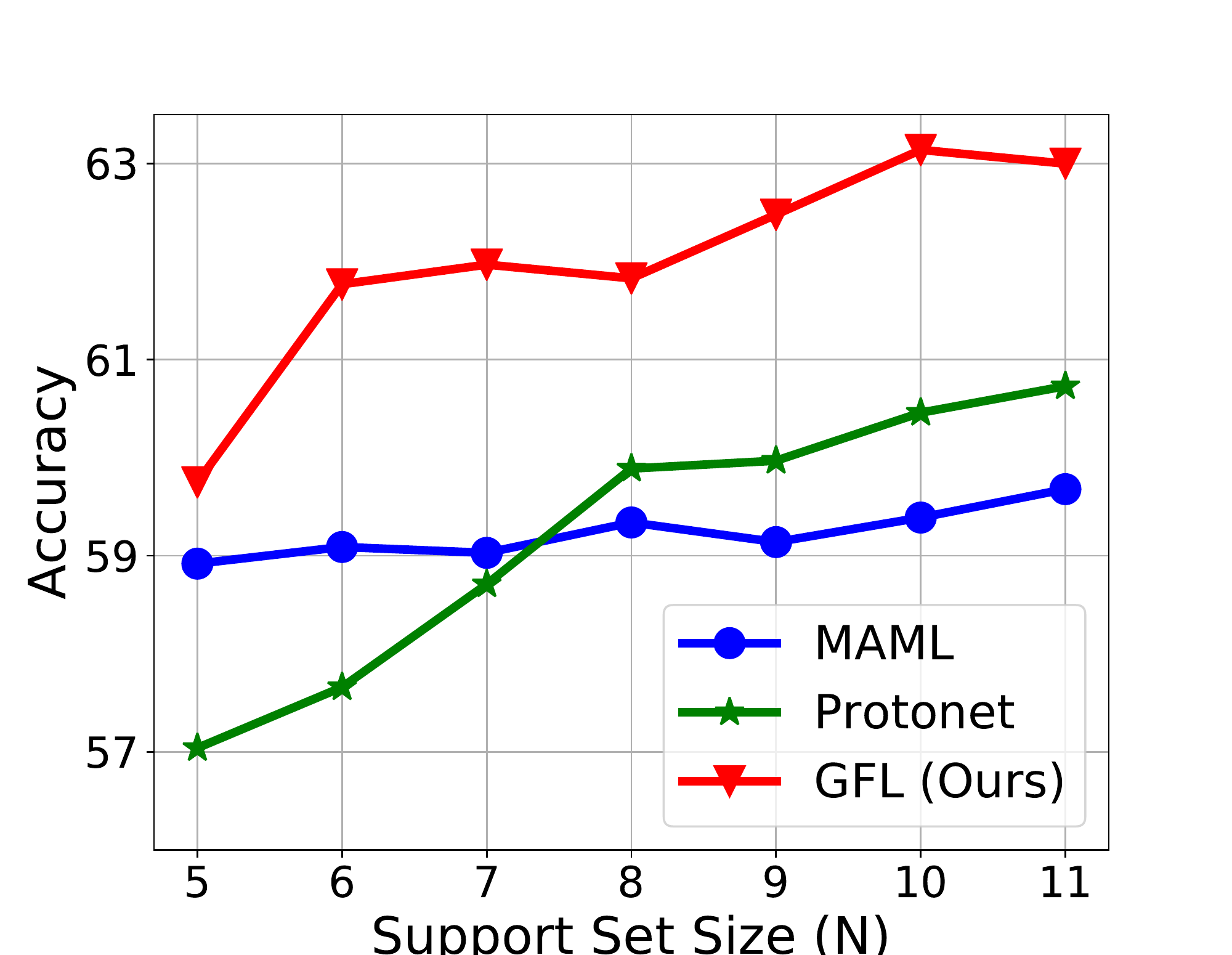}
		\caption{\label{fig:reddit_shot}: Reddit}
	\end{subfigure}
    \begin{subfigure}[b]{0.22\textwidth}
		\centering
		\includegraphics[height=0.8\textwidth]{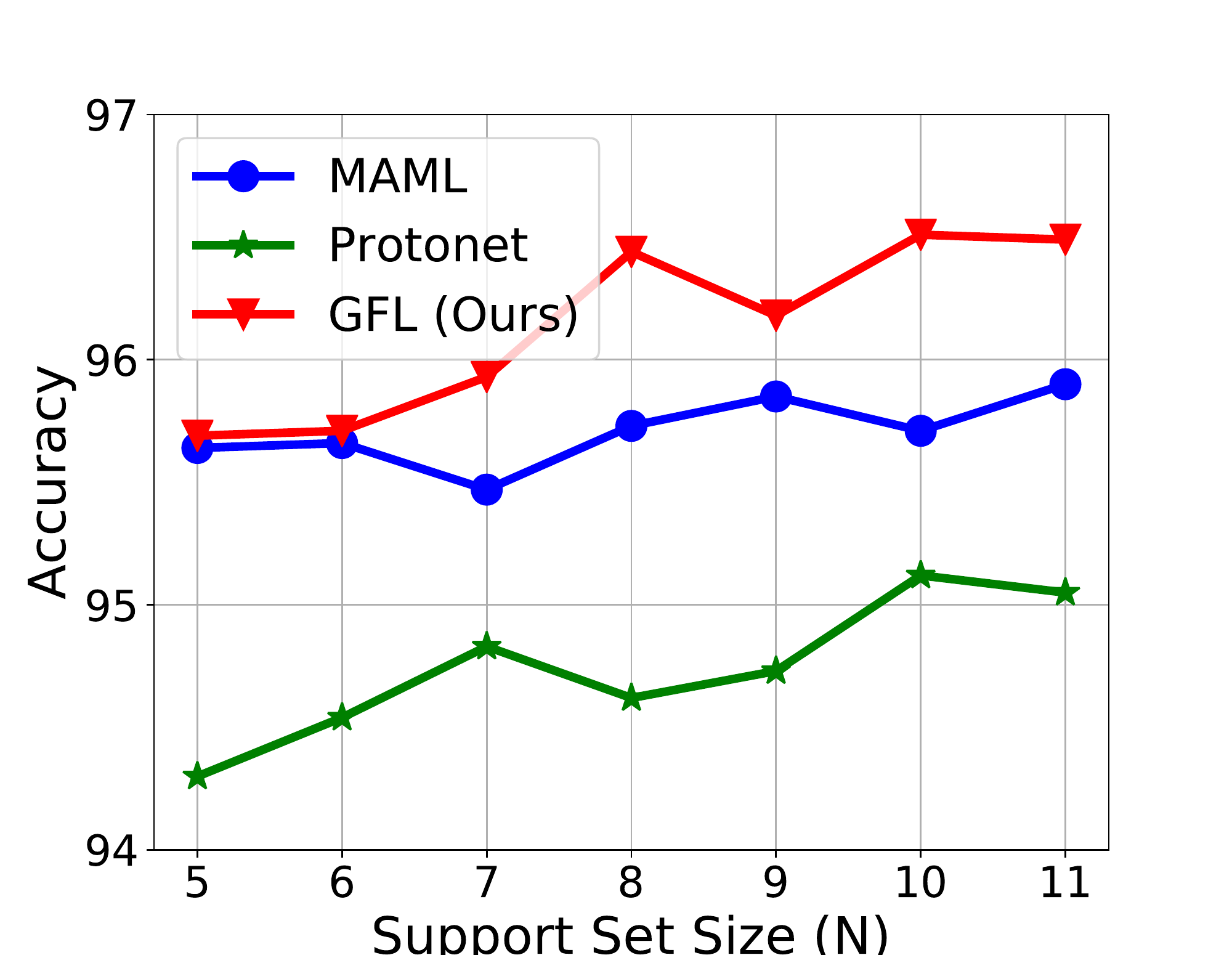}
		\caption{\label{fig:citation_shot}: Citation}
	\end{subfigure}
	\begin{subfigure}[b]{0.22\textwidth}
		\centering
		\includegraphics[height=0.8\textwidth]{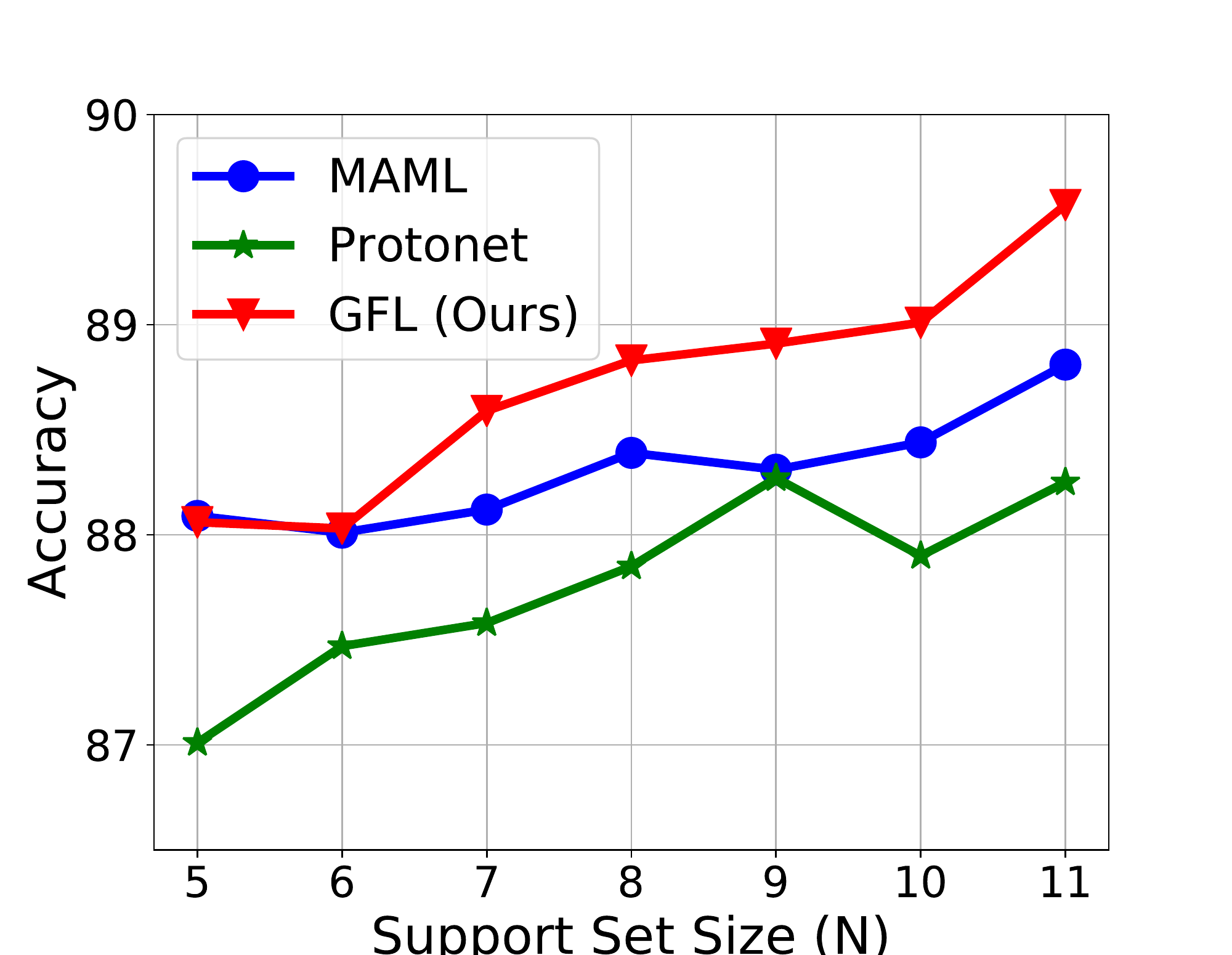}
		\caption{\label{fig:pubmed_shot}: Pubmed}
	\end{subfigure}
	\caption{Effect of support set size, which is represented by shot number $N$}
	\label{fig:illustration}
\end{figure}
\subsection{Sensitivity Analysis}
In this section, we analyze the sensitivities of the model to the size of the support set, threshold $\mu$ in the construction of the graph prototype, the similarity function for constructing relational structure $\mathcal{R}_i^k$, and different distance functions $d$.
\begin{figure*}[h]
	\centering
	\begin{subfigure}[c]{0.2\textwidth}
		\centering
		\includegraphics[height=29mm]{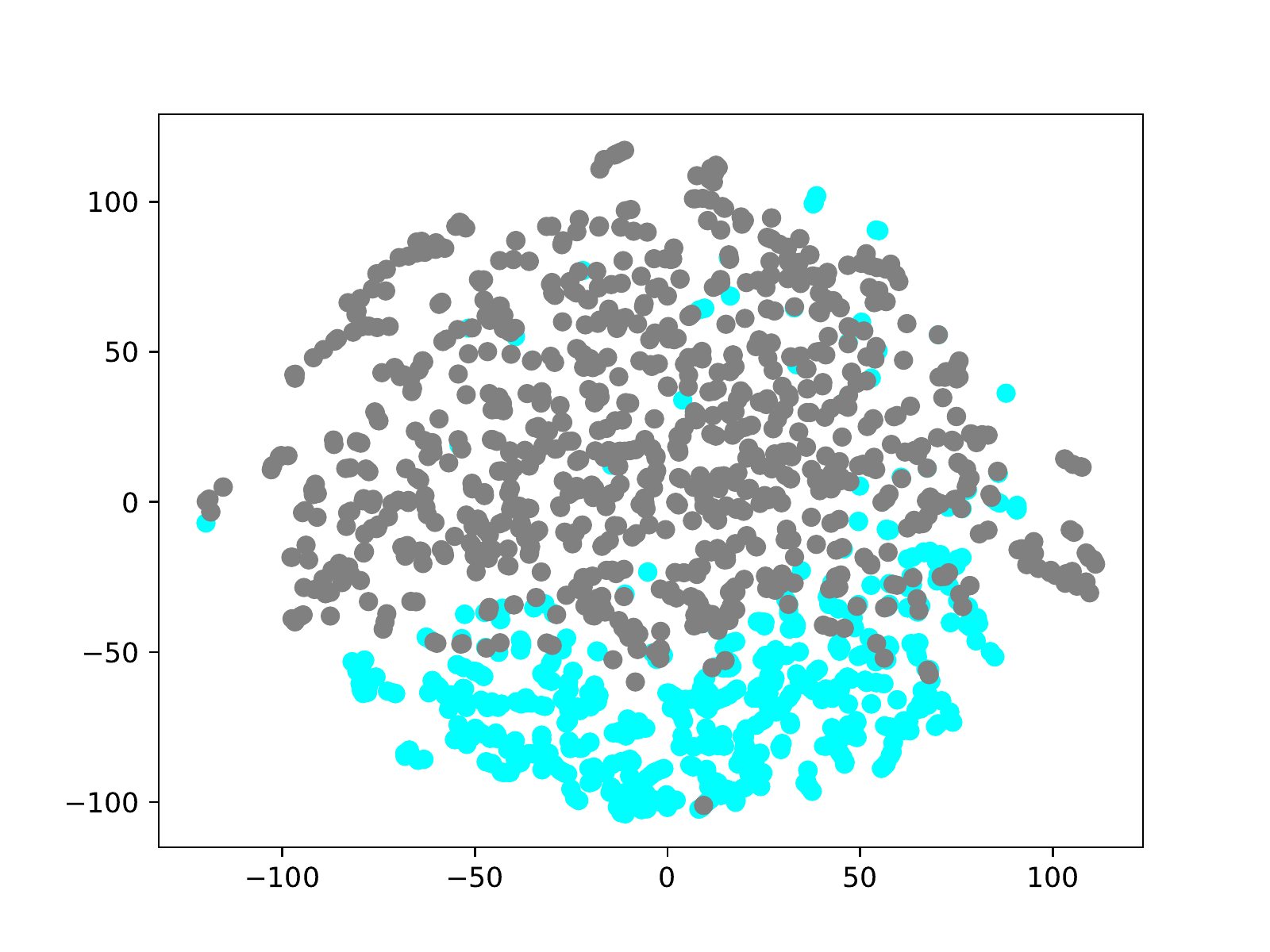}
		\caption{\label{fig:maml}: Class 1: GFL (Ours)}
	\end{subfigure}
	\begin{subfigure}[c]{0.2\textwidth}
		\centering
		\includegraphics[height=29mm]{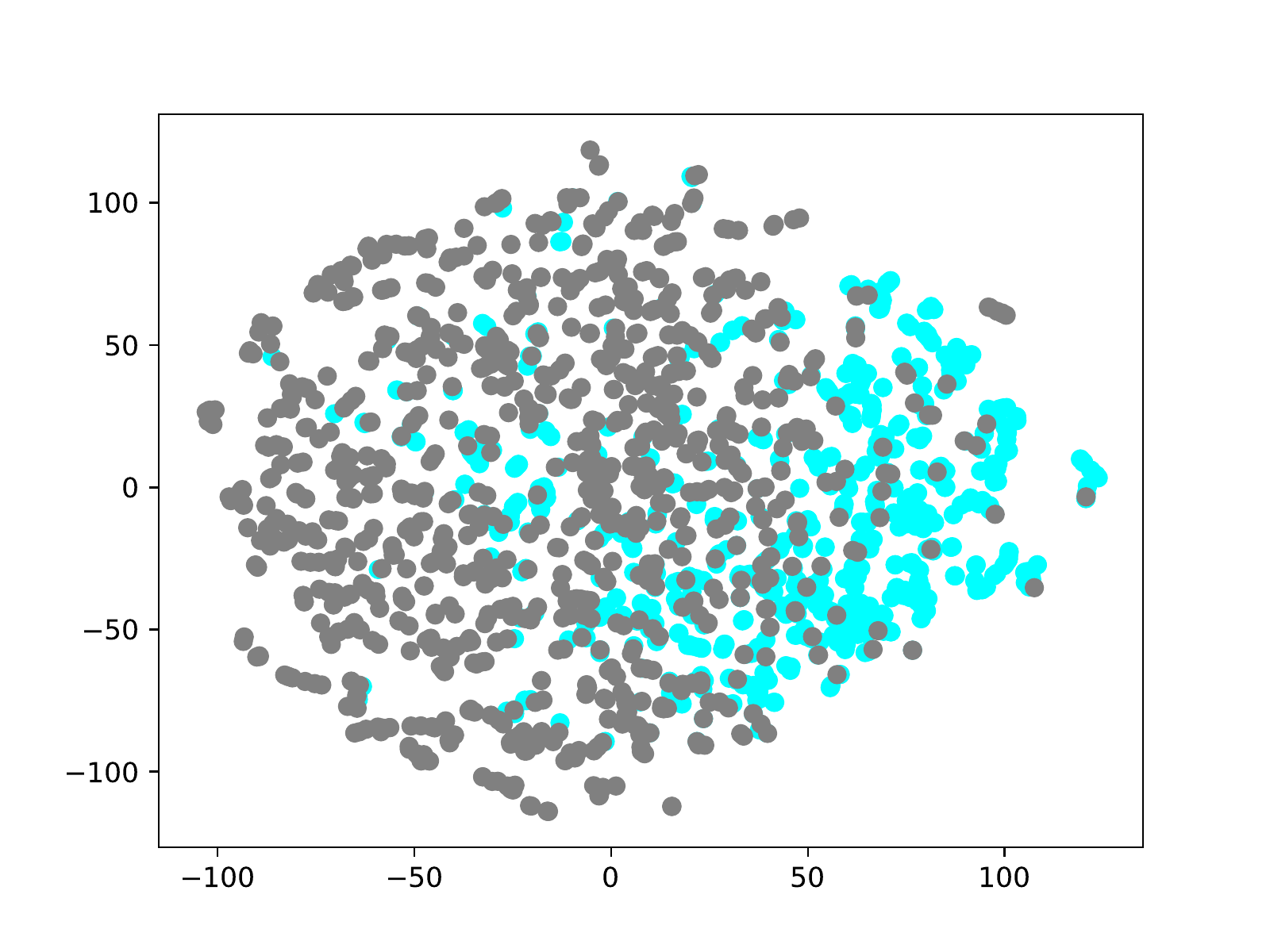}
		\caption{\label{fig:mumomaml}: Class 2: GFL (Ours)}
	\end{subfigure}
		\begin{subfigure}[c]{0.2\textwidth}
		\centering
		\includegraphics[height=29mm]{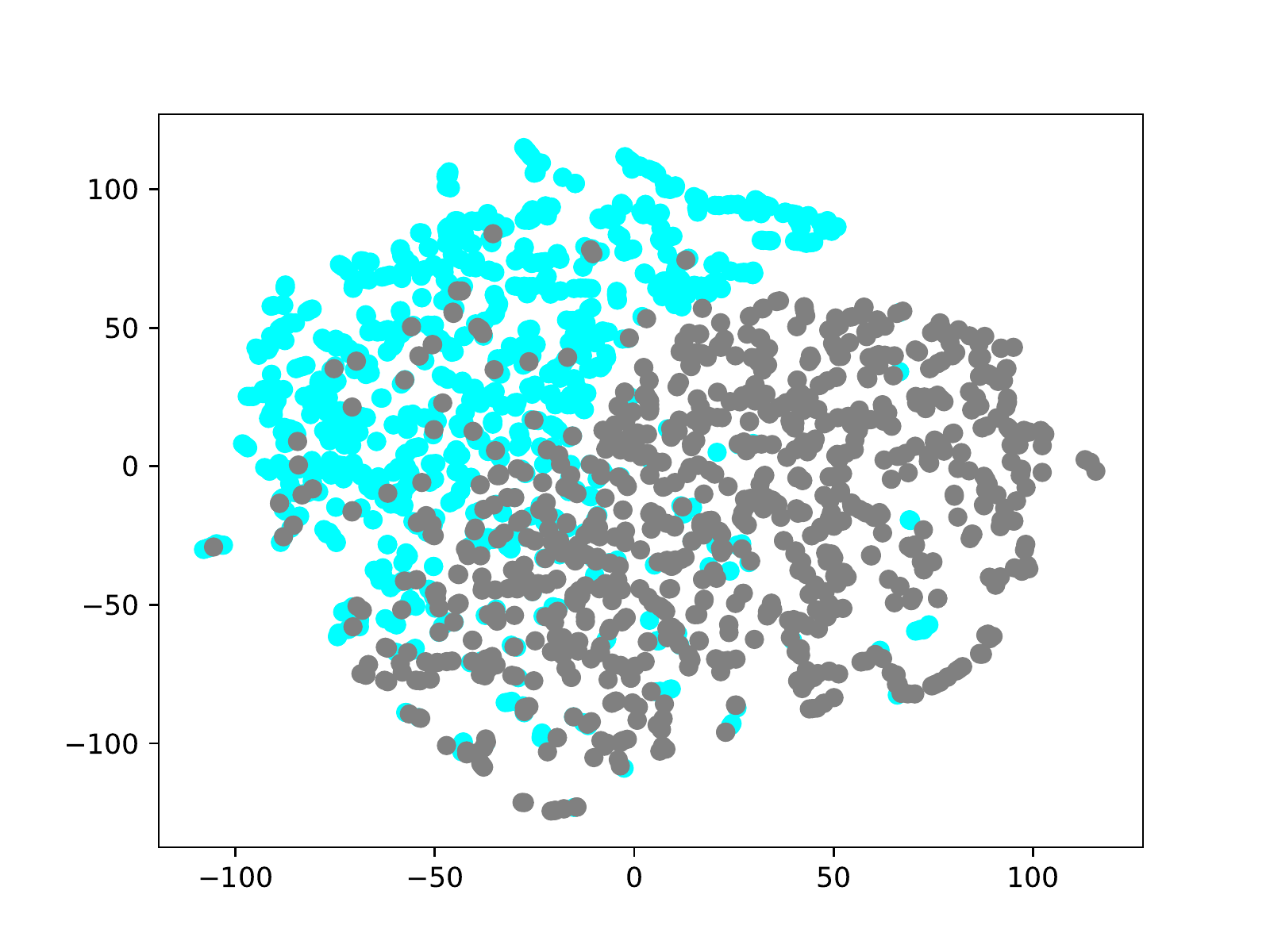}
		\caption{\label{fig:hsml}: Class 3: GFL (Ours)}
	\end{subfigure}
	\begin{subfigure}[c]{0.2\textwidth}
		\centering
		\includegraphics[height=29mm]{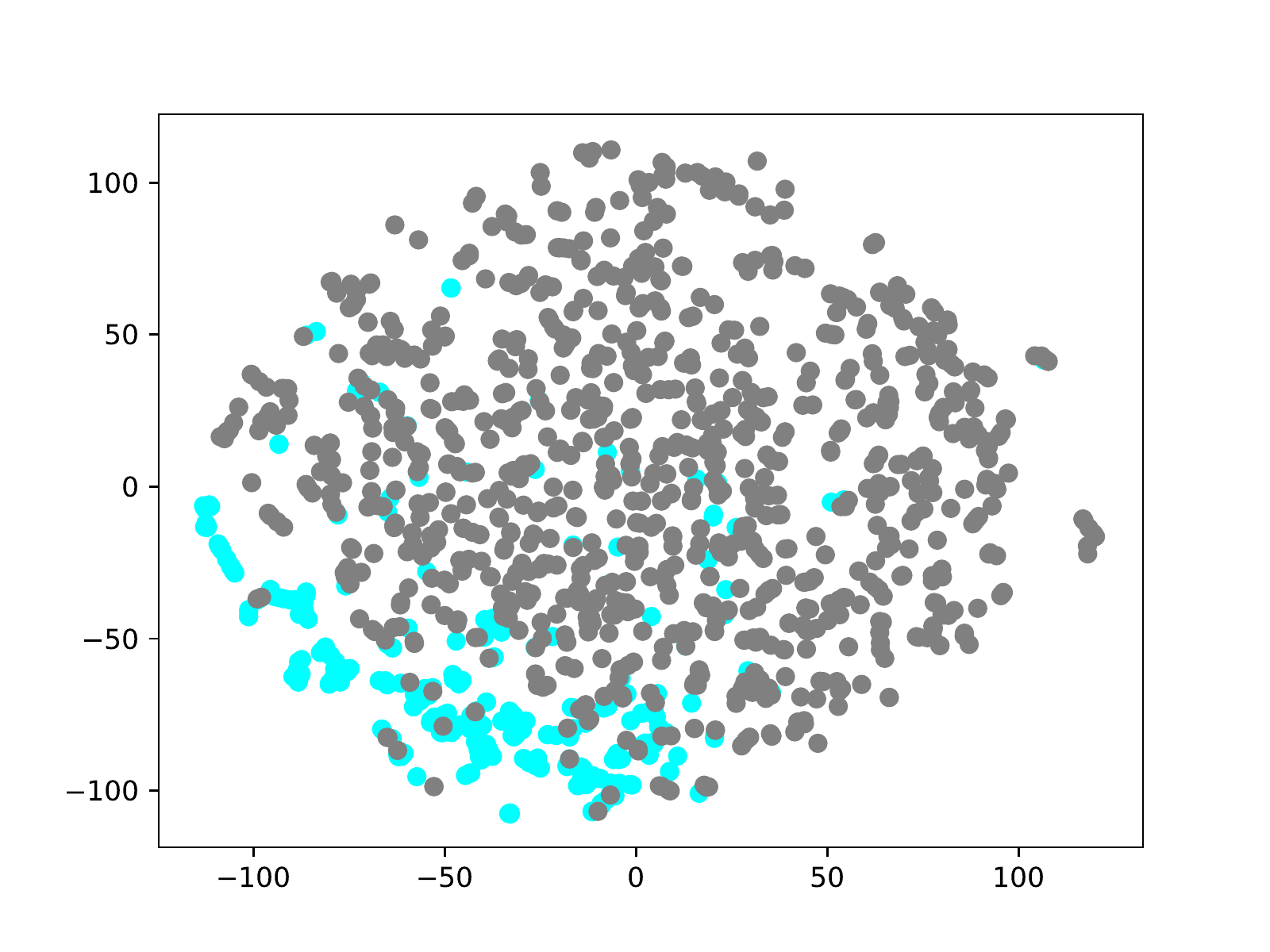}
		\caption{\label{fig:hsml}: Class 4: GFL (Ours)}
	\end{subfigure}
	\begin{subfigure}[c]{0.2\textwidth}
		\centering
		\includegraphics[height=29mm]{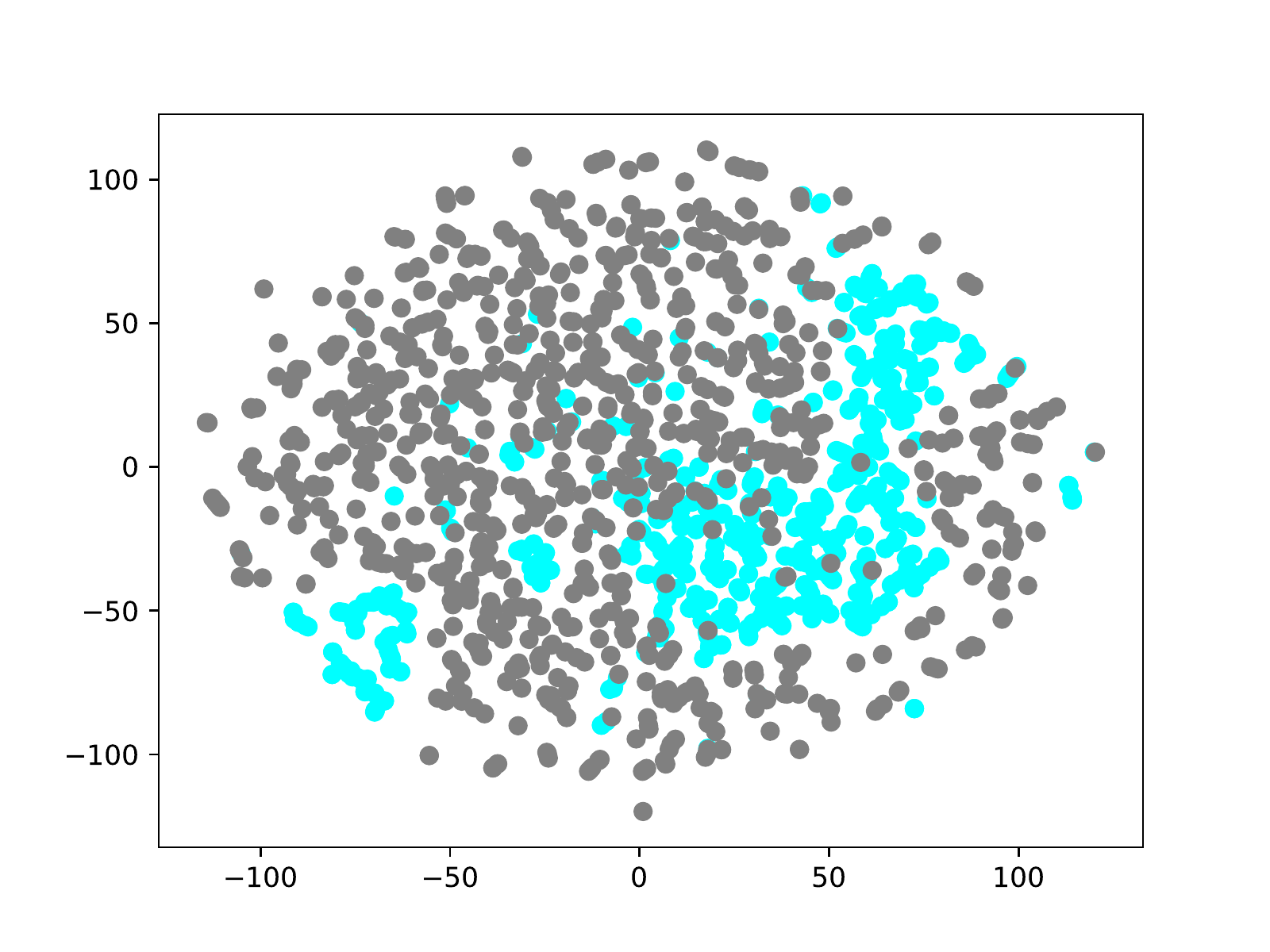}
		\caption{\label{fig:maml}: Class 1: Protonet}
	\end{subfigure}
	\begin{subfigure}[c]{0.2\textwidth}
		\centering
		\includegraphics[height=29mm]{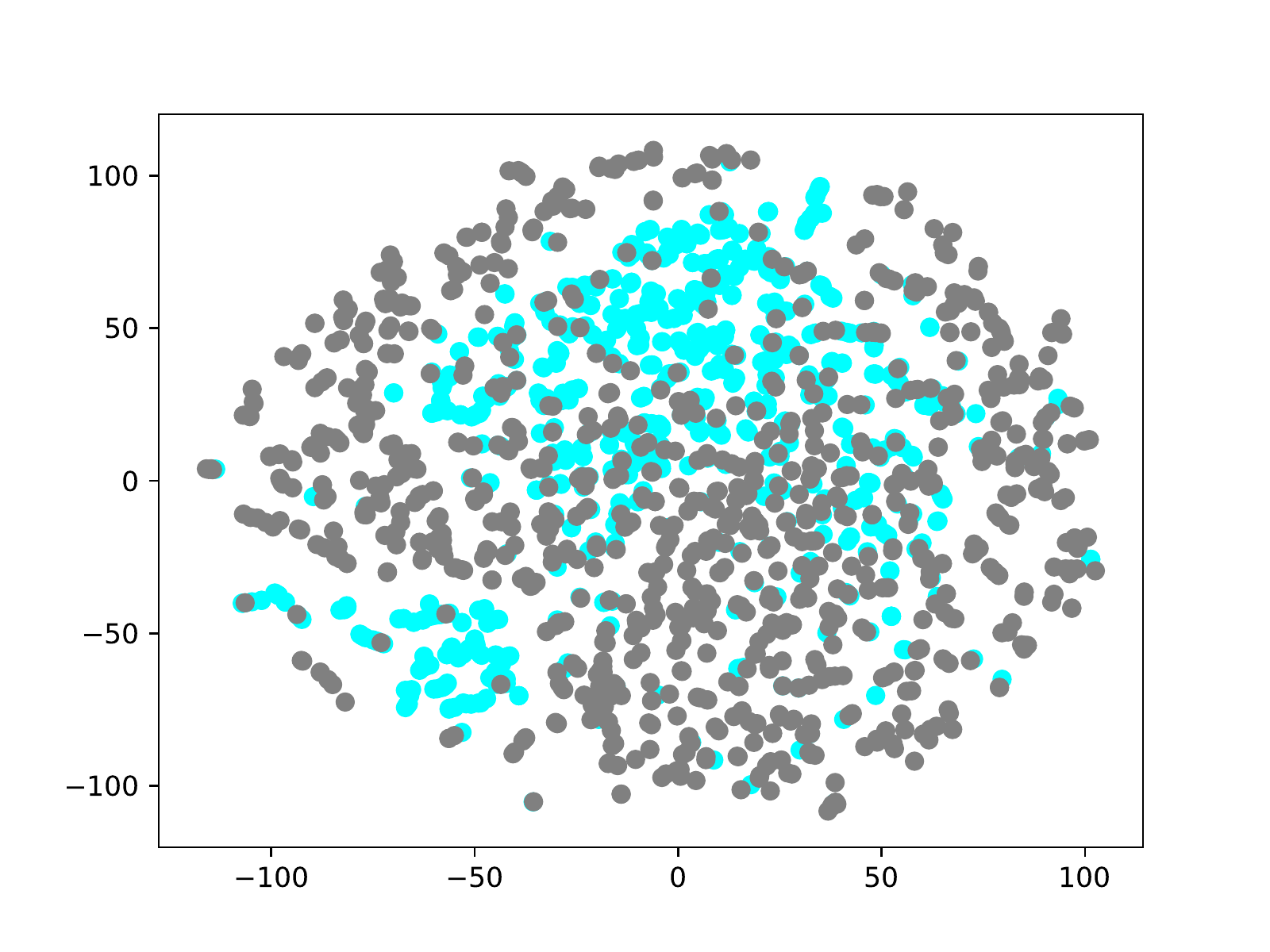}
		\caption{\label{fig:mumomaml}: Class 2: Protonet}
	\end{subfigure}
		\begin{subfigure}[c]{0.2\textwidth}
		\centering
		\includegraphics[height=29mm]{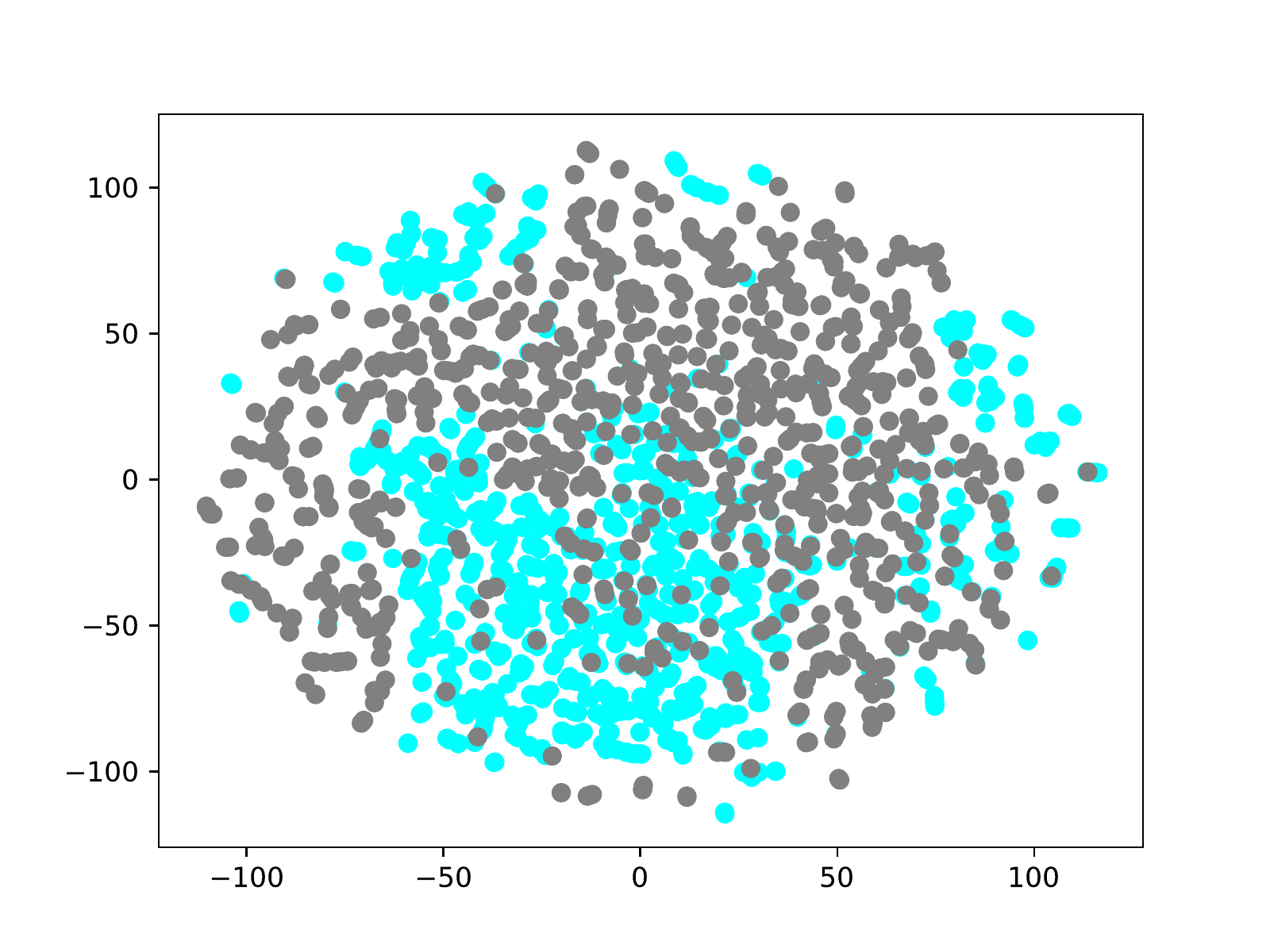}
		\caption{\label{fig:hsml}: Class 3: Protonet}
	\end{subfigure}
	\begin{subfigure}[c]{0.2\textwidth}
		\centering
		\includegraphics[height=29mm]{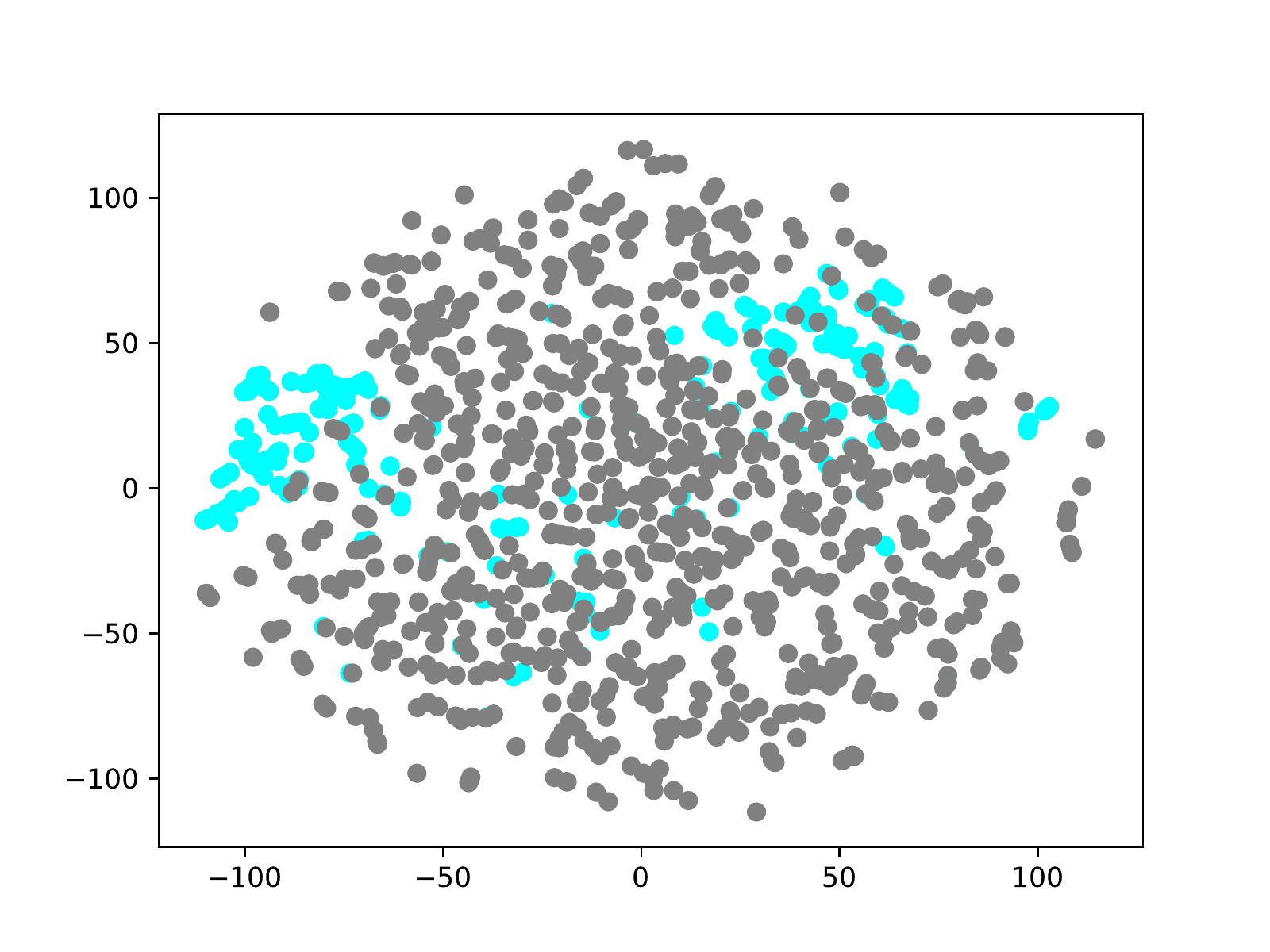}
		\caption{\label{fig:hsml}: Class 4: Protonet}
	\end{subfigure}
	\caption{Embedding visualization of positive (cyan nodes) and negative (grey nodes) data samples for each class.}
	\label{fig:illustration}
\end{figure*}
\subsubsection{Effect of Support Set Size}
We analyze the effect of the support set size, which is represented by the shot number $N$. For comparisons, we select two representative few-shot learning methods: Protonet (metric-learning based model) and MAML (gradient-based model). The results of each dataset are shown in Figure~\ref{fig:collaboration_shot}-\ref{fig:pubmed_shot}. We notice that when the support set size is small, Protonet performs worse than MAML. One potential reason could be that Protonet is sensitive to outliers, as it calculates prototype by averaging values over samples with equal weights. Thus, more data is expected to derive a reliable prototype. However, by extracting the relational structure among samples of the same class, our algorithm is more robust and achieves best performance in all scenarios.
\begin{table}[h]
\begin{center}
\small
\caption{Effect of threshold $\mu$ in relational structure construction of graph prototype. Results of Accuracy are reported.}
\label{tab:effect_threshold}
\begin{tabular}{l|c|c|c|c}
\toprule
$\mu$ & Collaboration & Reddit & Citation & Pubmed\\
\midrule
$0.5$ & $\mathbf{83.79\%}$ & $63.14\%$ & $\mathbf{96.51\%}$ & $89.37\%$\\
$0.6$ & $83.26\%$ & $\mathbf{63.43\%}$ & $96.19\%$ & $89.42\%$ \\
$0.7$ & $83.31\%$ & $63.17\%$ & $95.35\%$ & $\mathbf{89.60\%}$ \\
$0.8$ & $83.14\%$ & $63.21\%$ & $95.18\%$ & $89.21\%$ \\
\bottomrule
\end{tabular}
\end{center}
\end{table}

\begin{table}[h]
\begin{center}
\small
\caption{Effect of different similarity functions for calculating relational structure $\mathcal{R}_i^k$. Results of Accuracy are reported.}
\label{tab:effect_cn_method}
\begin{tabular}{l|c|c|c|c}
\toprule
Method & Collab. & Reddit & Cita. & Pubmed\\
\midrule
Jaccard & $82.98\%$ & $62.71\%$ & $95.18\%$ & $88.91\%$\\
Adamic-Adar & $83.70\%$ & $62.87\%$ & $95.49\%$ & $89.21\%$ \\
PageRank & $\mathbf{84.14\%}$ & $63.08\%$ & $95.93\%$ & $\mathbf{90.02\%}$ \\
Top-k CN & $83.79\%$ & $\mathbf{63.14\%}$ & $\mathbf{96.51\%}$ & $89.37\%$ \\
\bottomrule
\end{tabular}
\end{center}
\end{table}
\subsubsection{Effect of Threshold $\mu$ in Graph Prototype Construction} 
We further analyze the effect of threshold \begin{small}$\mu$\end{small} for constructing relational structure of the graph prototype. The results of \begin{small}$\mu=\{0.5,0.6,0.7,0.8\}$\end{small} are reported in Table~\ref{tab:effect_threshold}. In this table, the best threshold varies among different datasets. The effectiveness of the threshold demonstrates that the proposed model is robust to outliers. 

\subsubsection{Effect of Different Similarity Functions for Constructing Relational Structure $\mathcal{R}_i^k$} 
We further analyze the effect of different similarity functions for constructing relational structure of the graph prototype. Jaccard Index, Adamic-Adar~\cite{adamic2003friends}, PageRank and Top-k Common Neighbors (Top-k CN) are selected and the results are reported in Table~\ref{tab:effect_cn_method}. Note that, in previous results, we all use Top-k CN as similarity function. The results show
that GFL is not very sensitive to the similarity 
on a 
dataset 
may be achieved by different functions.
\subsubsection{Effect of Distance Functions $d$}
In addition, we replace the distance function $d$ in  Eqn. (2) from inner product (used in the original setting) for cosine distance. The results are reported in Table~\ref{app:tab_distance_function}. Compared with inner product, the similar results show that GFL is not very sensitivity to the distance function $d$. 
\begin{table}[h]
\begin{center}
\small
\caption{Effect of different distance functions $d$.}
\label{app:tab_distance_function}
\begin{tabular}{l|c|c|c|c}
\toprule
Method & Collab. & Reddit & Cita. & Pubmed\\
\midrule
GFL (inner product) & $83.79\%$ & $63.14\%$ & $96.51\%$ & $89.37\%$\\
GFL (cosine) & $84.02\%$ & $62.95\%$ & $96.02\%$ & $89.25\%$\\ \bottomrule
\end{tabular}
\end{center}
\end{table}

\subsection{Analysis of Learned Representation}
To better compare the learned representation of our model and Protonet, for each class, we use t-SNE~\cite{maaten2008visualizing} to visualize the embedding (i.e., \begin{small}$f_{\theta}$\end{small}) of positive samples (belong to this class) and 1000 negative samples (not belong to this class). The results of GFL and Protonet on collabration data are shown in Figure~\ref{fig:illustration}. Compared with Protonet, in this figure, we can see that our model can better distinguish the positive and negative samples.

\section{Conclusion}
In this paper, we introduce a new framework GFL to improve the effectiveness of semi-supervised node classification on a new target graph by transferring knowledge learned from auxiliary graphs. Built upon the metric-based few-shot learning, GFL integrates local node-level and global graph-level knowledge to learn a transferable metric space charaterized by node and prototype embedding functions. The empirical results demonstrate the effectiveness of our proposed model on four node classification datasets.
\section*{Acknowledgement}
The work was supported in part by NSF awards \#1652525, \#1618448, \#1849816, \#1925607, and \#1629914, the Army Research Laboratory under Cooperative Agreement Number W911NF-09-2-0053. The views and conclusions contained in this paper are those of the authors and should not be interpreted as representing any funding agencies.
\bibliographystyle{aaai}
{\fontsize{9.0pt}{10.0pt} \selectfont
\bibliography{ref}}
\appendix
\section{Additional Hyperparameter Settings}
\label{app:hyperparameter}
For more detailed hyperparameter settings, the learning rate is set as 0.01, the dimension of hierarchical graph representation $\mathbf{h}_i$ is set as 32. The reconstruction loss weight $\gamma$ is set as 1.0. Additionally, in order to construct the relational graph of few-shot labeled nodes for each class, we compute the similarity score between each two nodes by counting the number of k-hop ($k$=3) common neighbors and further smooth this similarity value by a sigmoid function. The computed similarity matrix is further fed into GNN for generating prototype embedding.
\section{Detailed Descriptions of Baselines}
\label{app:baseline}
We detail three types of baselines in this section. Note that, all GCN used in baselines are two-layers GCN with 32 neurons each layer. The descriptions are as follows:
\begin{itemize}[leftmargin=*]
    \item \textbf{Graph-based semi-supervised methods}:
    \begin{itemize}
        \item \textbf{Label Propagation (LP)}~\cite{zhu2002learning}: Label Propagation is a traditional semi-supervised learning methods.
        \item \textbf{Planetoid}~\cite{yang2016revisiting} Planetoid is a semi-supervised learning method based on graph embeddings. We use transductive formulation of Planetoid in this paper.
    \end{itemize}
    
    \item \textbf{Graph representation learning methods}:
    \begin{itemize}
        \item \textbf{Deepwalk}~\cite{perozzi2014deepwalk}: Deepwalk learns the node embedding in an unsupervised way. We concatenate the learned node embedding and the node features, then feed them to the multiclass classifier (using Scikit-Learn\footnote{https://scikit-learn.org/stable/}). Few-shot node labels in each graph are available for training classifer. 
        \item \textbf{node2vec}~\cite{grover2016node2vec}: This method is similar to the Deepwalk while we use node2vec model to learn node embedding. 
        \item 
        \textbf{Non-transfer-GCN}~\cite{kipf2016semi} In Non-transfer-GCN, we only train GCN on each meta-testing network without transferring knowledge from meta-training networks. 
    \end{itemize}
    \item \textbf{Transfer/Few-shot methods}
    \begin{itemize}
        \item \textbf{All-Graph-Finetune (AGF)} In AGF, we train GCN by feeding each meta-training graph one-by-one. Then, we can learn the initialization of GCN parameters from meta-training graphs. In meta-testing process, we finetune the learned initialization on every meta-testing graphs. The hyperparameters (e.g., learning rate) of AGF are the same as GFL.
        \item \textbf{K-nearest-neighbor} (K-NN) Similar as the settings of~\cite{triantafillou2019meta}, we first learn the initialization of GCN parameters by using all meta-training graphs. Then, in the testing process, we use the learned embedding function (i.e., GCN) to learn the representation of support nodes and each query node. Finally, we use k-NN for classification.
        \item \textbf{Matching Network (Matchingnet)}~\cite{vinyals2016matching}: Matching network is a metric-based few-shot learning method. Since traditional matching network focuses on one-shot learning, in our scenario, we still use each query node representation to match the most similar on in support set  and use its label as the predicted label for the query node.
        \item \textbf{MAML}~\cite{finn2017model}: MAML is a representative gradient-based meta-learning method, which learn a well-generalized model initialization which can be adapted with a few gradient steps. For MAML, we set the learning rate of inner loop as 0.001.
        \item 
        \textbf{Prototypical Network (Protonet)}~\cite{snell2017prototypical} Prototypical network is a representative metric-based few-shot learning method, which constructs the prototype by aggregating nodes belong to the same class using mean pooling.
    \end{itemize}
\end{itemize}
\end{document}